\documentclass[preprint,12pt,authoryear]{elsarticle}

\usepackage{amssymb}
\usepackage{amsmath}
\usepackage{booktabs}
\usepackage{multirow}
\usepackage{multicol}
\usepackage{comment}
\usepackage{acronym}
\usepackage{setspace}
\usepackage{caption}
\usepackage{subcaption}
\usepackage{array}

\newacro{CNN}[CNN]{convolutional neural network}
\newacro{MAPE}[MAPE]{mean absolute percentage error}
\newacro{MDAPE}[MdAPE]{median absolute percentage error}
\newacro{MAE}[MAE]{mean absolute error}
\newacro{MSE}[MSE]{mean square error}
\newacro{RMSE}[RMSE]{root mean square error}

\journal{Ecological Informatics}

\begin{document}

\begin{frontmatter}



\title{Computer vision-based estimation of small arthropod biomass}


\author[syke]{Mikko Impiö\corref{cor1}} 
\ead{mikko.impio@syke.fi}
\author[due,zwu]{Philipp M. Rehsen} 
\author[au]{Jarrett Blair} 
\author[au]{Cecilie Mielec} 
\author[due,zwu]{Arne J. Beermann} 
\author[due,zwu]{Florian Leese} 
\author[au]{Toke T. Høye\fnref{last}} 
\author[jyu]{Jenni Raitoharju\fnref{last}} 
\cortext[cor1]{Corresponding author}
\fntext[last]{Joint last author}

\affiliation[syke]{organization={Finnish Environment Institute (Syke)},
            city={Helsinki},
            country={Finland}}
\affiliation[due]{organization={University of Duisburg-Essen, Aquatic Ecosystem Research},
            city={Essen},
            country={Germany}}
\affiliation[zwu]{organization={University of Duisburg-Essen, Centre for Water and Environmental Research (ZWU)},
            city={Essen},
            country={Germany}}
\affiliation[au]{organization={Aarhus University, Department of Ecoscience},
            city={Aarhus},
            country={Denmark}}
\affiliation[jyu]{organization={University of Jyväskylä, Faculty of Information Technology},
            city={Jyväskylä},
            country={Finland}}

\begin{abstract}
The ability to estimate invertebrate biomass using only images could help scaling up quantitative biodiversity monitoring efforts.
Computer vision-based methods have the potential to omit the manual, time-consuming, and destructive process of dry weighing specimens.
We present two approaches for dry mass estimation that do not require additional manual effort apart from imaging the specimens: fitting a linear model with novel predictors, automatically calculated by an imaging device, and training a family of end-to-end deep neural networks for the task, using single-view, multi-view, and metadata-aware architectures.
We propose using area and sinking speed as predictors. These can be calculated with BIODISCOVER, which is a dual-camera system that captures image sequences of specimens sinking in an ethanol column.
For this study, we collected a large dataset of dry mass measurement and image sequence pairs to train and evaluate models.
We show that our methods can estimate specimen dry mass even with complex and visually diverse specimen morphologies.
Combined with automatic taxonomic classification, our approach achieves a median individual error of 10-20\% and produces group-level dry mass distributions that do not differ significantly from observed distributions.
We highlight the importance of choosing appropriate evaluation metrics, and encourage using both percentage errors and absolute errors as metrics, because they measure different properties.
We also explore different optimization losses, data augmentation methods, and model architectures for training deep-learning models.
\end{abstract}




\end{frontmatter}



\section{Introduction}\label{sec-introduction}
Reversing biodiversity loss requires a deep understanding of natural ecosystems and their ongoing changes \citep{ipbes2019Summary,commission2021EU}.
This need is particularly apparent for many invertebrate taxa, which remain understudied compared to vertebrates and other charismatic species \citep{ceballos2017Biological}.
Invertebrates are integral in ecosystem processes and provide important ecosystem services, such as pollination, food provisioning, and recycling of organic matter \citep{eisenhauer2023Ecosystem, hallmann2017More, noriega2018Research}. 

Collecting information on invertebrate biomass and abundances is important, as it gives us information on ecosystem processes, energy flow, and prey-predator relationships \citep{wardhaugh2013Estimation,ganihar1997Biomass,benke1999LengthMass}.
A direct way to infer specimen biomass is weighing them.
Usually this means measuring or estimating the dry mass, which involves a time-consuming drying process and is destructive for the specimen.
Non-destructive biomass estimation, where dry-weighing is not needed, usually relies on heuristic regression models based on body measurements, such as length and width \citep{schoener1980LengthWeight,benke1999LengthMass,ganihar1997Biomass,rogers1976General}, or using wet mass of a bulk sample as a proxy measurement for biomass.
Regression models predicting biomass from specimen length are by far the most popular specimen-level estimation method \citep{wardhaugh2013Estimation}, and many studies are dedicated to calculating regressions for specific taxonomic groups \citep{benke1999LengthMass,gruner2003Regressions,wardhaugh2013Estimation,ganihar1997Biomass,rogers1976General}. Wet and dry mass are strongly correlated, and conversion factors between the two have also been calculated for many taxonomic groups \citep{gogina2022Weighttoweight, hu2022Measuring}. Wet or fresh weight highly depends on the taxonomic group, used preservative and storing time \citep{wetzel2005Preservation}.
Other specimen-level approaches include approximation based on fitting geometric shapes on the specimen \citep{llopis-belenguer2018Evaluation, wardhaugh2013Estimation} and using specimen width as an additional predictor \citep{wardhaugh2013Estimation}.

Regardless, bulk wet weight measurements do not provide information on the size distribution of specimens within a sample. For these fine-grained analyses, the manual effort required for measuring or drying is prohibitive for scaling up monitoring efforts. Processing large amounts of samples produced in monitoring programs can be a costly and time-consuming bottleneck \citep{karlsson2020Swedish}.
Molecular methods, such as DNA metabarcoding, can make large-scale species diversity estimation easier.
While some studies suggest that abundance estimation from sequencing read abundances can be feasible for some taxonomic groups \citep{iwaszkiewicz-eggebrecht2026Accuracy, sickel2023Abundance}, this kind of abundance estimation still suffers from taxon-specific biases, calling for further research \citep{iwaszkiewicz-eggebrecht2026Accuracy, creedy2019Accurate}. 
One suggested way to improve abundance estimates at bulk sample level is combining read abundances with wet weight measurements \citep{iwaszkiewicz-eggebrecht2023FAVIS}.
Automated biomass estimation approaches that do not rely on expert knowledge or tedious handling of specimens are needed to efficiently estimate the biomass of a large number of field samples.


Some automatic invertebrate biomass estimation methods have been proposed, using various approaches ranging from community-level estimation using radar \citep{wotton2019Mass,huppop2019Perspectives} to the more common approach of specimen-level biomass estimation using images \citep{schneider2022Bulk,arje2020Automatic}.
The latter is more effective, as biomass and abundance can be estimated separately for different functional groups when combined with automatic identification.
Existing specimen-level approaches rely on measuring the area of specimens in the images and fitting a linear model to estimate the dry mass \citep{schneider2022Bulk,arje2020Automatic}.
\citet{arje2020Automatic} calculated a mixed linear model between the species area (measured in pixels) and the dry mass.
They used the BIODISCOVER imaging device, the same device as used in this study.
As BIODISCOVER captures several images for each specimen, Ärje et al.~used the average area of each specimen to fit the linear model.

\citet{schneider2022Bulk} used a similar approach where approximately 20 specimens per functional group were used to calculate a per-pixel density for the group.
This per-pixel density was then used to estimate biomass for new specimens based on their area.
Their approach used a top-down camera, which captures a single image from a bulk sample, and they cropped each specimen with a watershed algorithm.
A similar top-down imaging device was proposed by \citet{wuhrl2022DiversityScanner}, which was used by \citet{shirali2026Automated} for estimating the area of different insect body parts (head, thorax, abdomen) with computer vision methods. No biomass estimation was done in this study, but with a sufficient dataset of image-mass pairs, this approach of separating body parts could produce useful input features for machine learning systems and biomass estimation.

\citet{marstaller2019Deepbees} used a \ac{CNN} to calculate body proportions of honeybees.
They did not estimate biomass, but automatic body proportion calculation could produce biomass estimates using traditional methods, such as those mentioned above.
\citet{gu2026Automating} used a deep learning system to estimate crab carapace proportions for biomass estimation.

In this study, we present a large dataset of dry mass measurements for individual specimens, with their associated images and other metadata. We approach the biomass estimation problem from two angles: a) traditional linear modeling, but using features that are made possible with an imaging device, such as the BIODISCOVER, which captures multi-view image sequences of specimens, and b) state-of-the-art computer vision approaches with neural network models, including multi-view and metadata-aware models.

For linear modeling, we use a new, useful proxy for mass: the sinking speed of the specimen.
The BIODISCOVER device works by capturing a series of images of specimens as they sink through an ethanol-filled cuvette. This means that the frame rate of the BIODISCOVER’s cameras can be combined with the specimen’s distance traveled between frames to measure the specimen’s sinking velocity. Given that the sinking velocity of an object is affected by its density, combining a specimen’s sinking velocity with its area could be an effective method for estimating its mass \citep{walker2021Estimation}. 

The use of neural networks, particularly \acp{CNN}, is motivated by the fact that not all specimen area corresponds to the same amount of biomass.
Different body parts have different densities.
For example, a dipteran wing can take up a large area in an image, but its biomass is only a small percentage of the total.
Accurate estimation should take into account these different visual features and their effect on total specimen biomass.
Deep learning models are known to be useful for learning useful representations of complex inputs and can in theory learn mappings between visual features and biomass, given enough training data.

We present different modeling approaches and evaluate them using  a novel dataset of 1116 individually imaged and dry-weighed specimens.
We compare different architecture choices using different evaluation metrics. We also discuss the importance of choosing an appropriate metric for both neural network optimization and evaluation of models.



\section{Materials and methods}\label{materials-and-methods}
\subsection{Datasets}
\label{sec-data} 

We collected two arthropod image datasets for this study, including both terrestrial and aquatic species.
The first dataset consists of 980 specimens, collected and imaged at Aarhus University in Denmark, and the second dataset 136 specimens collected at the Finnish Environment Institute in Finland.
The first dataset was morphologically identified to the taxonomic level of order, and the second to the level of species, later referred to as the \textit{Order dataset} and the \textit{Species dataset}. 
The Order dataset was collected in 2020 from emergence traps in agricultural habitats in Denmark, and imaged and weighed during 2021.
The Species dataset was collected in fall of 2021, as part of a routine monitoring program targeting benthic species in lakes in Finland, and imaged and weighed in 2023.
The Species dataset specimens were stored in 96\% ethanol prior to imaging, while the Order dataset used 75\% ethanol.

\begin{figure}[ht]
     \centering
     \begin{subfigure}{0.2\textwidth}
        \centering
        \includegraphics[width=\textwidth]{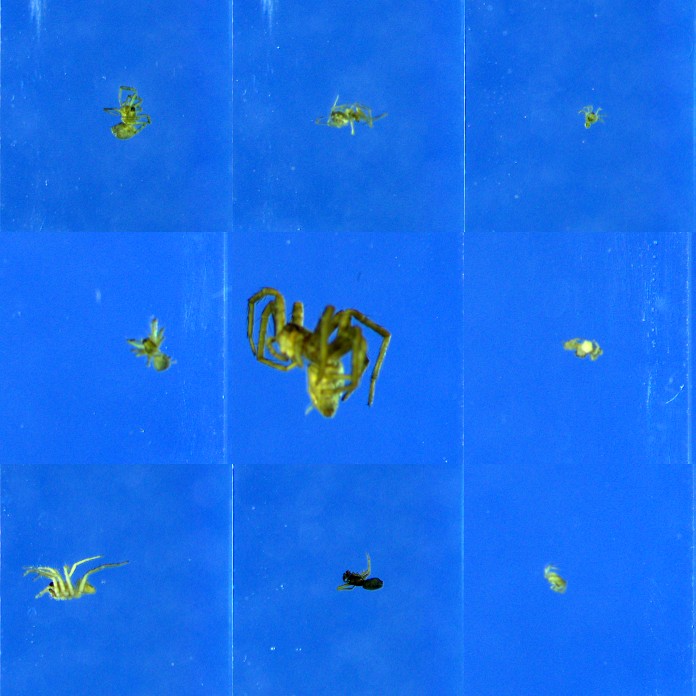}
        \caption{Araneae (O)}
     \end{subfigure}
     \quad
     \begin{subfigure}{0.2\textwidth}
         \centering
        \includegraphics[width=\textwidth]{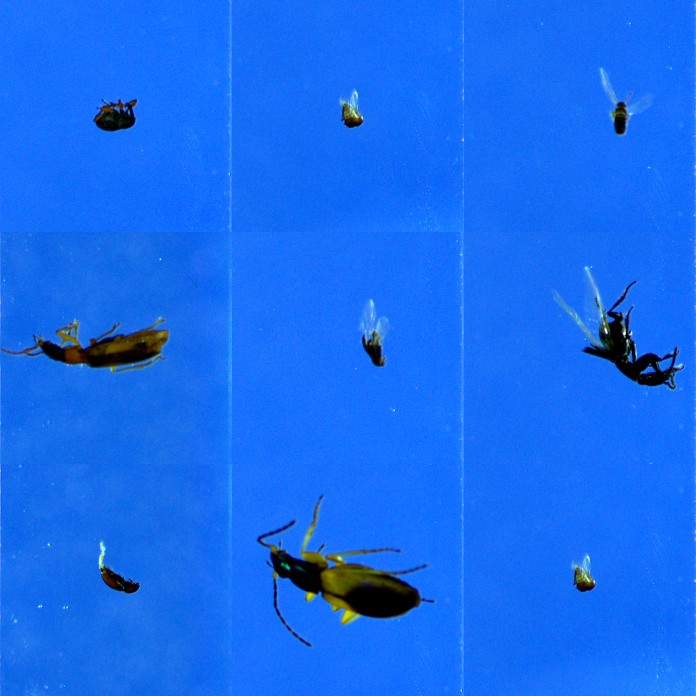}
        \caption{Coleoptera (O)}
     \end{subfigure}
     \quad
     \begin{subfigure}{0.2\textwidth}
        \centering
        \includegraphics[width=\textwidth]{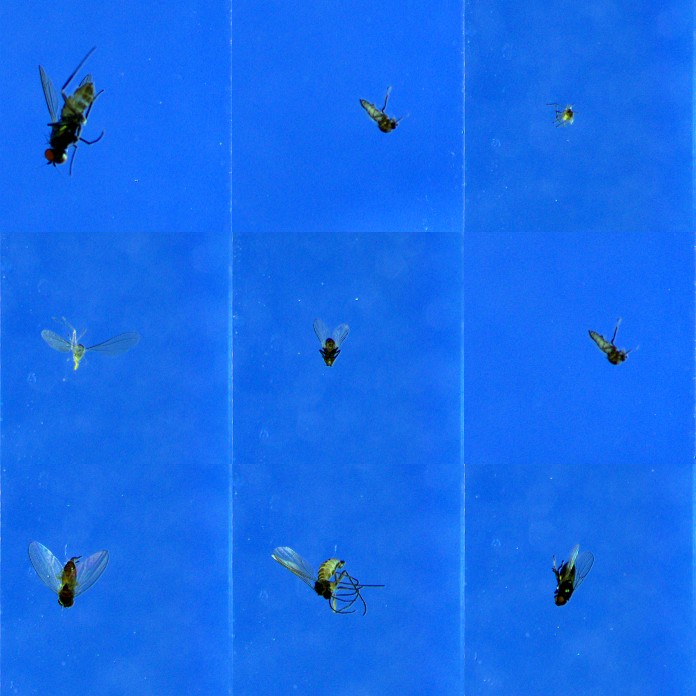}
        \caption{Diptera (O)}
     \end{subfigure}
     \\
     \begin{subfigure}{0.2\textwidth}
        \centering
        \includegraphics[width=\textwidth]{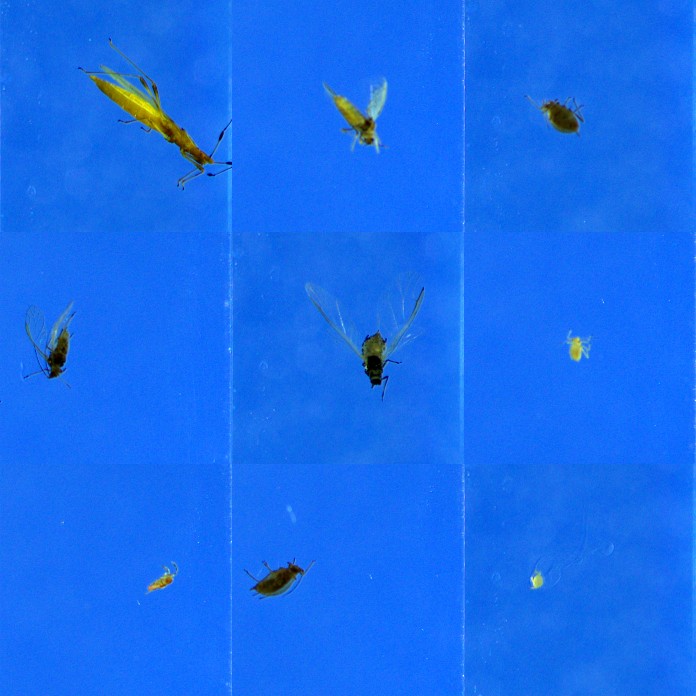}
        \caption{Hemiptera (O)}
     \end{subfigure}
     \quad
     \begin{subfigure}{0.2\textwidth}
        \centering
        \includegraphics[width=\textwidth]{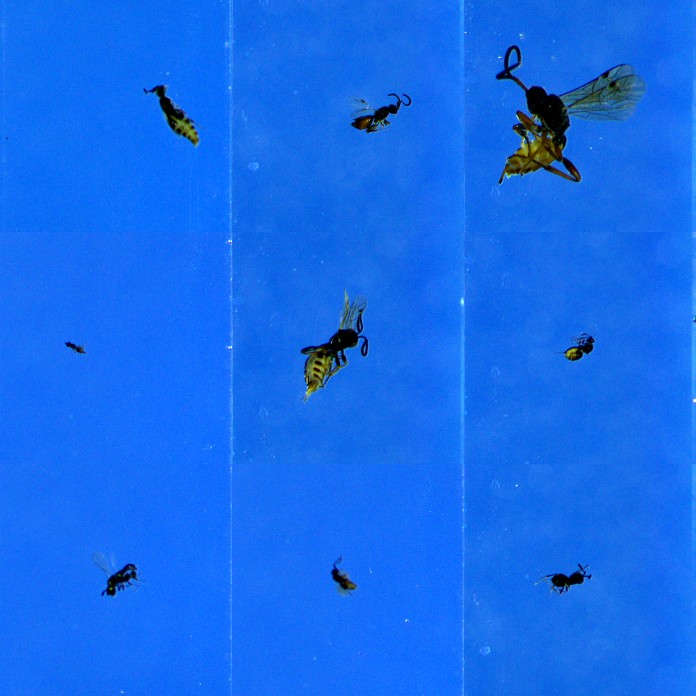}
        \caption{Hymenoptera (O)}
     \end{subfigure}
     \\
     \begin{subfigure}{0.2\textwidth}
        \centering
        \includegraphics[width=\textwidth]{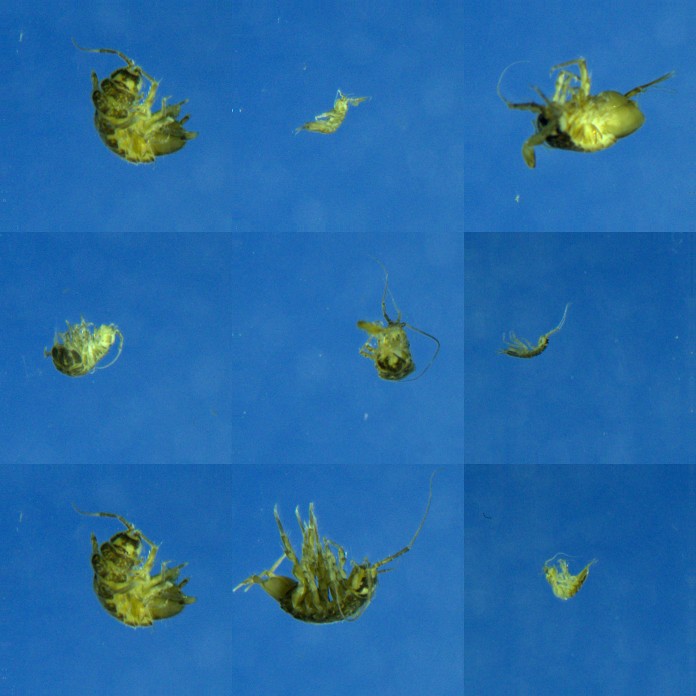}
        \caption{Asellus aquaticus (S)}
     \end{subfigure}
     \quad
     \begin{subfigure}{0.2\textwidth}
        \centering
        \includegraphics[width=\textwidth]{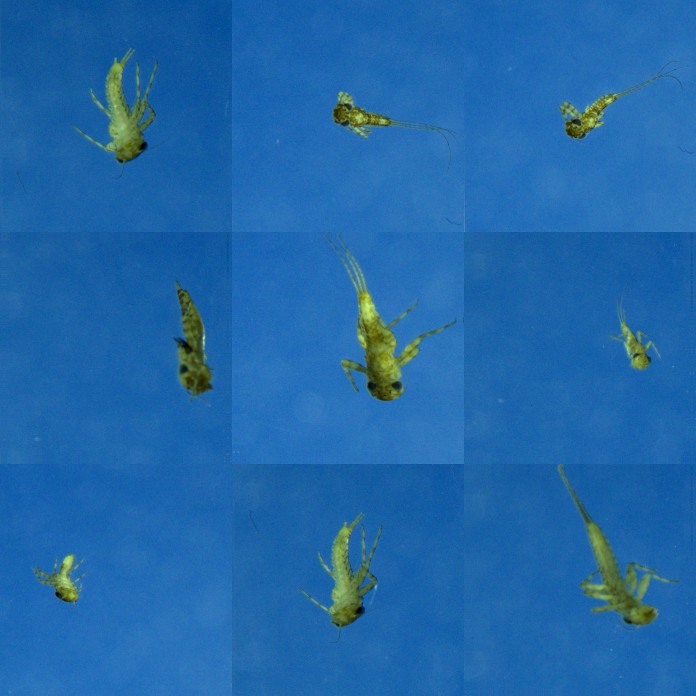}
        \caption{Kageronia fuscogrisea (S)}
     \end{subfigure}
     \quad
     \begin{subfigure}{0.2\textwidth}
        \centering
        \includegraphics[width=\textwidth]{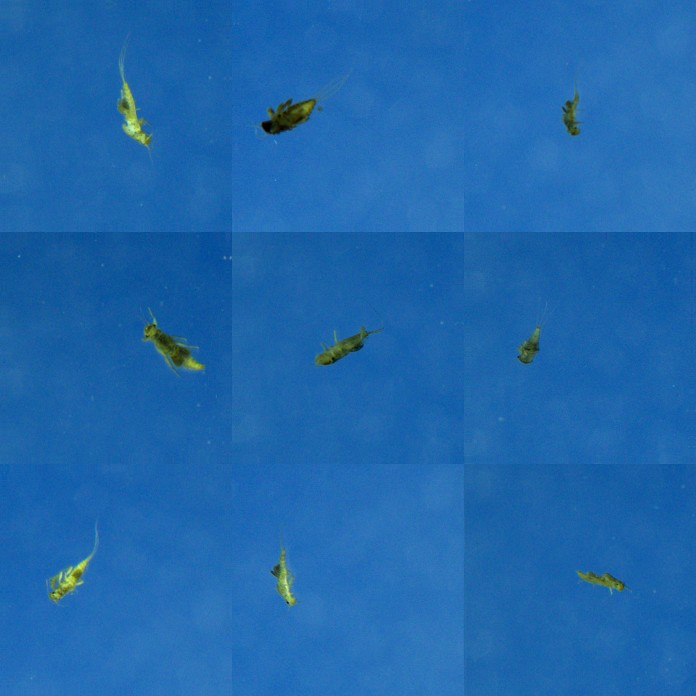}
        \caption{Caenis horaria (S)}
     \end{subfigure}
\caption{Randomly sampled images from both datasets grouped by taxonomic order. There is large variation in size and biomass within groups. S=Species dataset, O=Order dataset. The Species dataset was imaged completely separately, using a different device from the Order dataset, thus providing a test dataset for out-of-distribution generalization.}
\label{fig-images}
\end{figure}

The taxa present in the Order dataset are Araneae, Coleoptera, Diptera, Hemiptera, and Hymenoptera.
The taxa present in the Species dataset are the isopod \textit{Asellus aquaticus}, and the mayflies \textit{Caenis Horaria} and \textit{Kageronia fuscogrisea}, chosen because they are among the most common taxa in Finnish benthic surveys.
Examples image frames can be seen in Figure~\ref{fig-images}.

\begin{figure} [t]
    \centering 
    \includegraphics[width=0.7\textwidth]{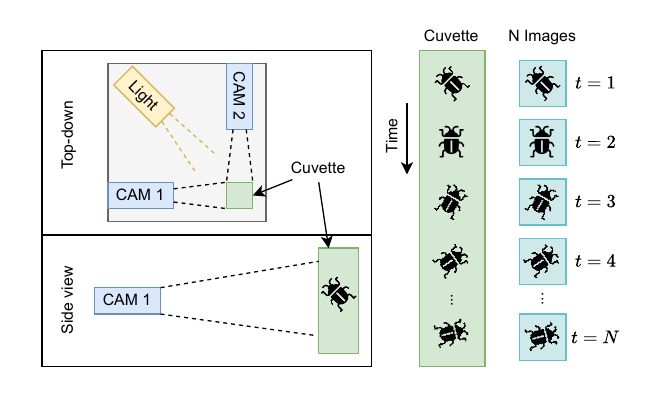}
    \caption{\label{fig-biodiscover-overview}Overview of the BIODISCOVER imaging device setup. The device images specimens from two angles as they fall through an ethanol-filled cuvette. Useful metadata, such as sinking speed, can be calculated from the image timestamps, locations, and total number of images, as the camera frame rate is fixed.}
\end{figure}

Both datasets were imaged with a BIODISCOVER device (described in detail by \citet{arje2020Automatic}), which produces image sequences from two angles.
The device has two cameras at a 90-degree angle photographing a 1 cm $\times$ 1 cm $\times$ 3.5 cm cuvette, where a specimen is dropped by a human operator.
An overview of the imaging setup can be seen in Figure~\ref{fig-biodiscover-overview}.
The sinking specimen is photographed by both cameras, and the image sequence is saved on disk with accompanying metadata.
The metadata includes the area of the specimen, the position of the frame in the cuvette, and frame rate information.
The area is automatically calculated by the BIODISCOVER software, by comparing image frames to a calibration image \citep{arje2020Automatic}.
Since the frame rate is fixed and all specimens travel the same distance across the imaging area, specimen sinking time is proportional to the number of frames captured per specimen. A low number of recorded images means that the specimen took less time to sink through the cuvette. Conversely, more images means that the same distance took more time, i.e., a lower sinking speed.
The Order and Species datasets were imaged in completely separate settings with different copies of the BIODISCOVER device.

Our dataset has only square images, but the size of the saved image depends on specimen size.
The Order dataset image resolution is always $464 \times 464$ pixels.
The Species dataset was collected with an updated version of the BIODISCOVER software that has the same pixel density, but uses a tighter crop of $448 \times 448$ for smaller specimens.
To keep the scale compatible, we padded the size of these images to match the $464 \times 464$ images, by mirroring the border by 8 pixels in all four directions.
By visual inspection, this does not produce artifacts to the images, or changes to specimen areas, as specimens are usually in the middle of the image frame.
Square images retain scale information without distortions when images are resized for the neural network input.

The sinking speed depends on the ethanol used to fill the cuvette, and the ethanol used to store the specimens.
We used denatured 91\% ethanol during imaging of the Species dataset and 75\% ethanol for the Order dataset. Theoretically, this causes a slight difference in the falling speeds across the datasets, as the ethanol concentrations during storing and imaging are slightly different (96\% to 91\% for Species dataset, 75\% to 75\% for Order dataset.

After imaging, the specimens were dried in a drying oven.
For the Order dataset, the specimens were dried for 72 hours at 50 degrees Celsius, and weighed straight after being taken out of the oven with a precision of $1.0~\mu g$
For the Species dataset, the specimens were dried for 22-24 hours in 105 degrees Celsius.
These specimens were placed in a vacuum desiccator to prevent them from collecting moisture before weighing.
The weighing was done using a precision of $0.5~\mu g$.

An overview of the datasets and their statistics can be seen in Table~\ref{tbl-dataset-details}. The biomass statistics include mean, median, maximum, and minimum biomass in micrograms, as well as the 25th and 75th percentiles, and the standard deviation of the group.
Successful training of machine learning models depends on high-quality datasets with enough training data.
This has previously been a challenge in image-based biomass estimation, as there are only few publicly available datasets with image-data pairs. We compare our dataset to previous datasets found in the literature in Table~\ref{tbl-dataset-comparison}.

\begin{figure}[ht]
\centering
\begin{subfigure}{0.4\textwidth}
    \includegraphics[width=\textwidth]{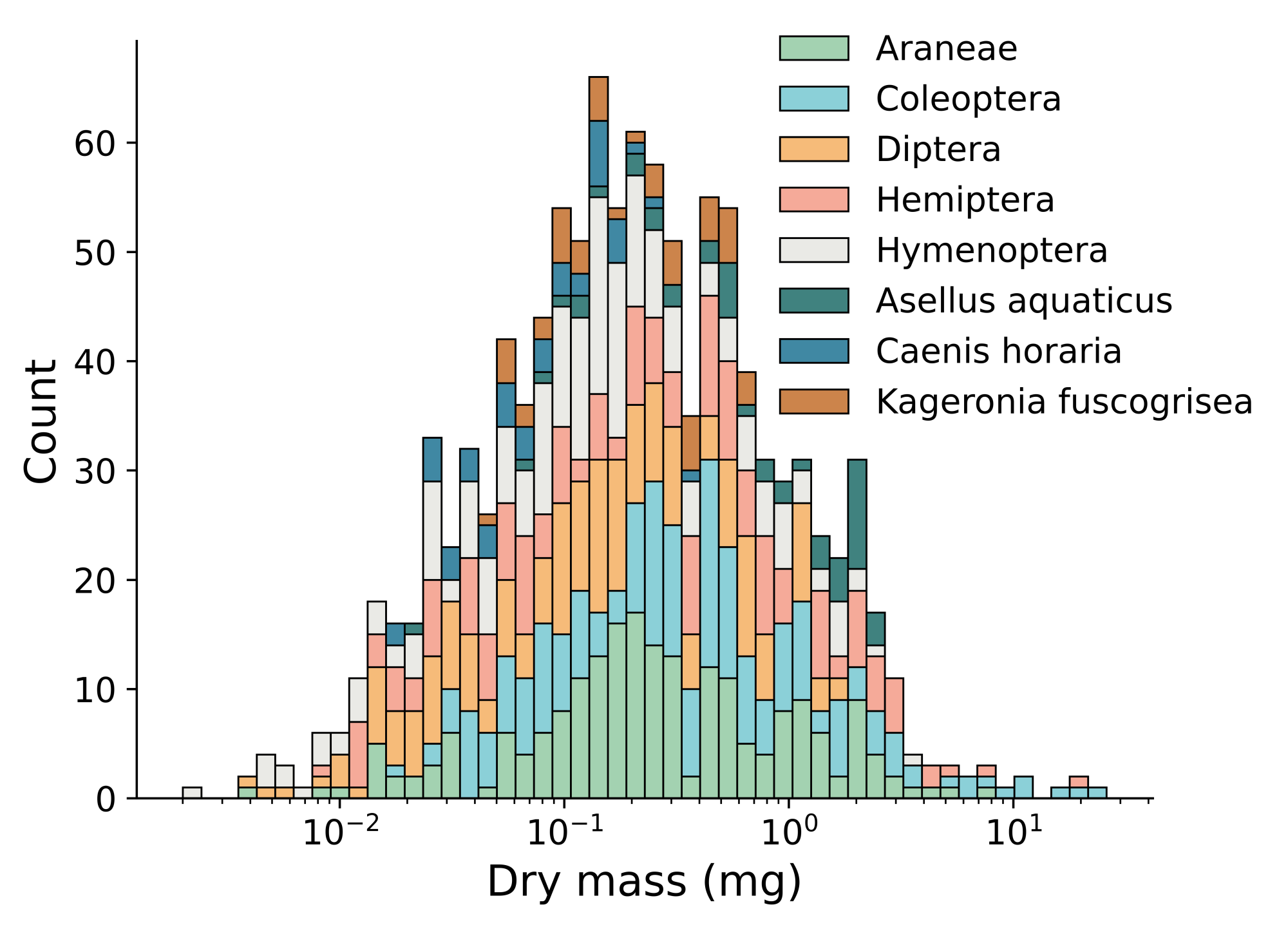}
    \caption{The mass distributions for both datasets.}
    \label{fig-weight-distribution}
\end{subfigure}
\quad
\begin{subfigure}{0.4\textwidth}
    \includegraphics[width=\textwidth]{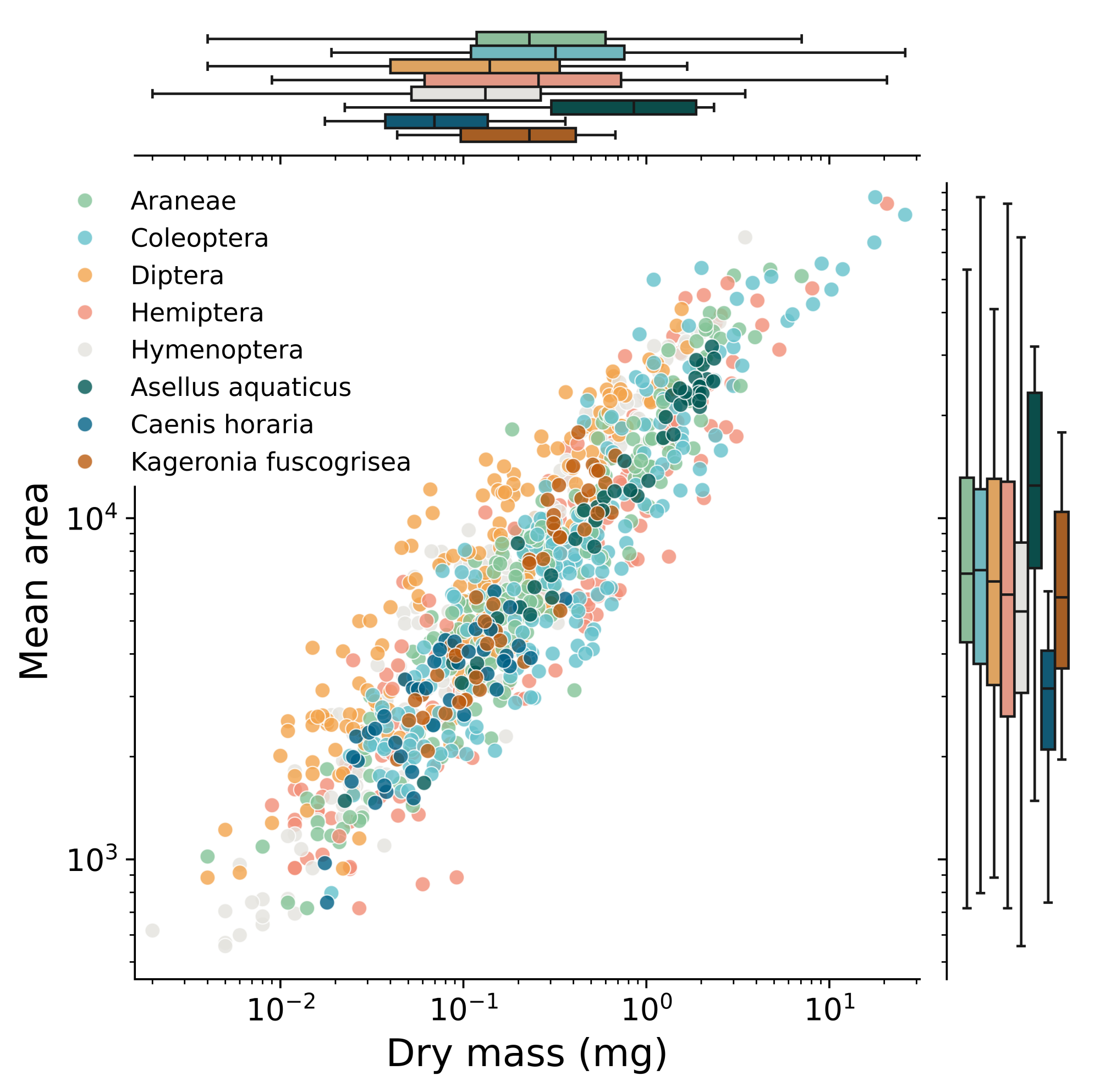}
    \caption{Relationship of biomass and specimen mean area.}
    \label{fig-weight-vs-area}
\end{subfigure}
\\
\begin{subfigure}{0.4\textwidth}
    \includegraphics[width=\textwidth]{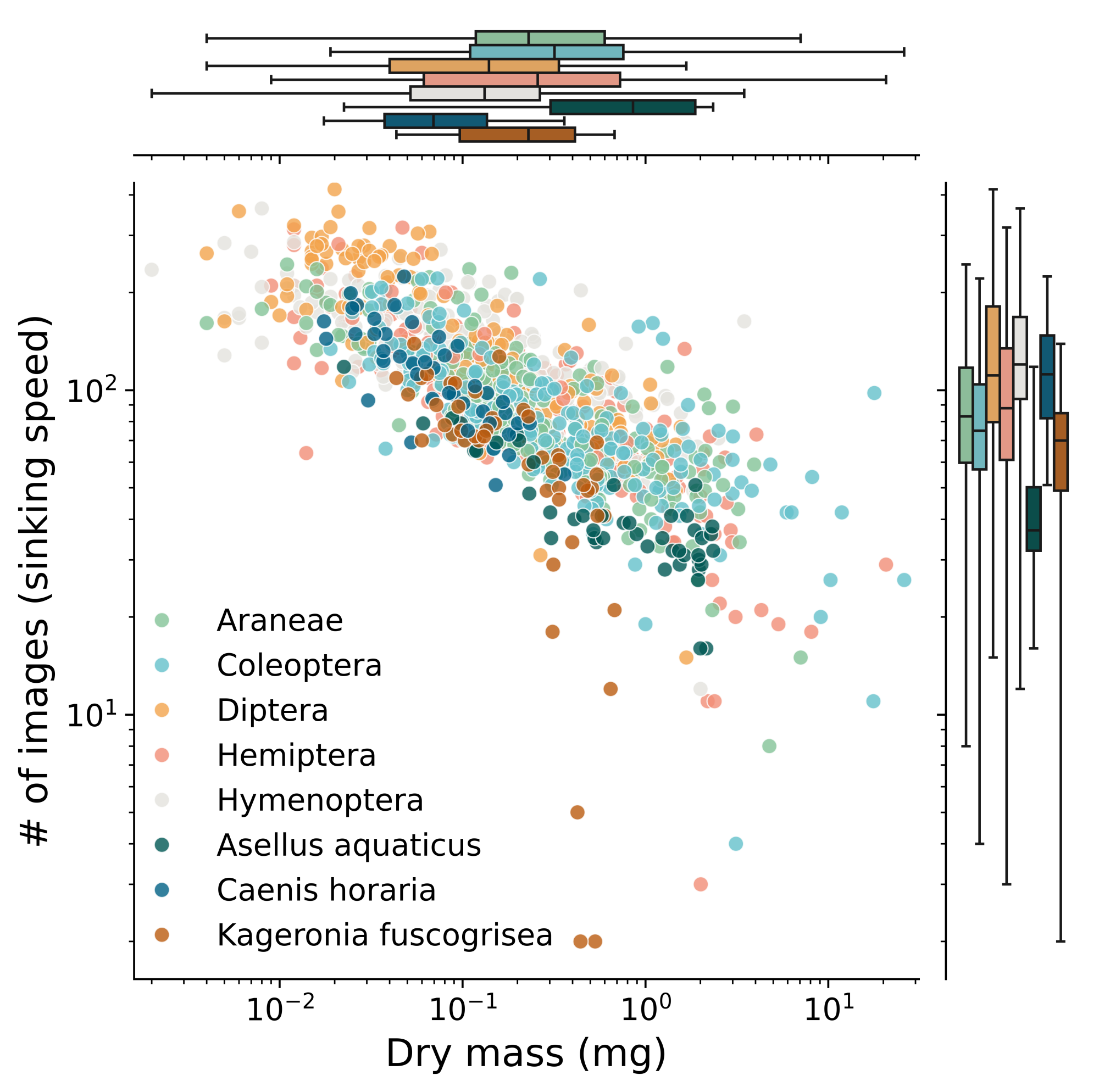}
    \caption{Relationship of biomass and image count, which is a proxy for sinking speed.}
    \label{fig-weight-vs-images}
\end{subfigure}
\quad
\begin{subfigure}{0.4\textwidth}
    \includegraphics[width=\textwidth]{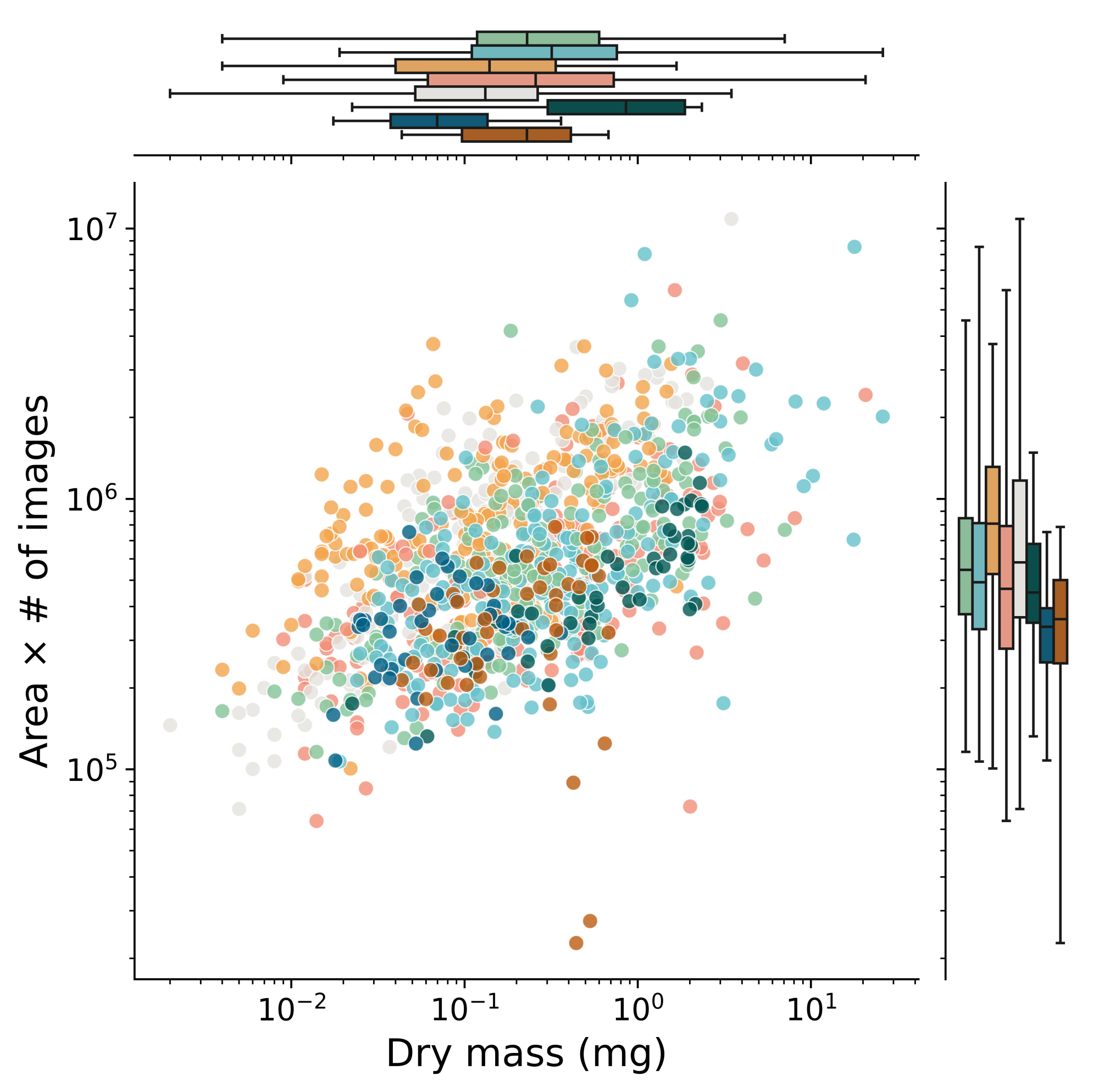}
    \caption{Relationship of biomass and product of image count and mean area.}
    \label{fig-weight-vs-mass}
\end{subfigure}
\caption{Histogram of dry mass measurements (a), and relationships between mass measurements and dataset features (b-d).}
\label{fig-dataset-stats}
\end{figure}

The biomass distributions in our datasets are shown in Figure~\ref{fig-weight-distribution}.
The distributions are close to a log-normal distribution with truncation in high and low values.
This is due to practical issues: too large specimens do not fit in the BIODISCOVER device, and too small specimens are difficult to weigh after drying.
The specimen dry mass is highly correlated with specimen mean area and image count, as shown in Figures~\ref{fig-weight-vs-area} and \ref{fig-weight-vs-images}.
A pseudo-mass feature, derived from the product of specimen mean area and image count, is displayed in Figure~\ref{fig-weight-vs-mass}.

\begin{figure}[ht]
    \centering
    \includegraphics[width=\textwidth]{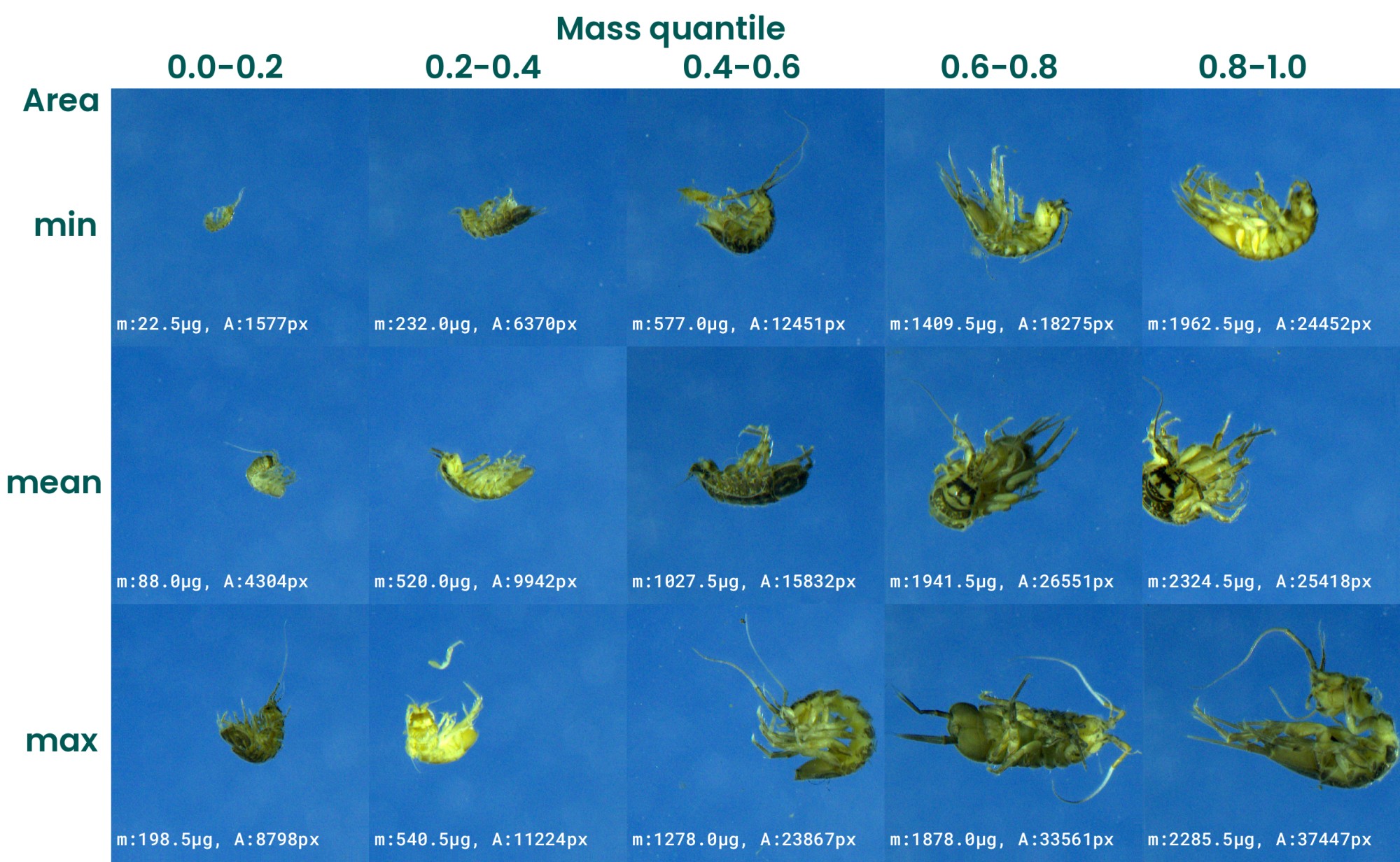}
    \caption{Area and dry mass ranges of \textit{Asellus aquaticus} specimens. The x-axis shows samples from five mass quantiles (0-20\%, 20\%-40\%, 40\%-60\%, 60\%-80\%, and 80\%-100\%) and the y-axis shows minimum, mean and maximum area from these quantiles. Below each image, the specimen mass (m) and area (A) of the specimen in this particular image is given.}
    \label{fig-weight-area-example}
\end{figure}

Figure~\ref{fig-weight-area-example} shows an example of the range of area and dry mass for a single species, \textit{Asellus aquaticus} in the Species dataset.
The images illustrate how specimens with similar areas might have very different dry masses.
For example, the specimen with the maximum area in the mass quantile 0.6-0.8 has a mass of $1878~\mu g$, while a smaller specimen, closest to the mean area, is heavier, with a mass of $1941~\mu g$.

\begin{table}[ht]
\centering
\caption{Overview of our datasets. Dry masses (m) are presented in micrograms (µg). 25p=25th percentile, 75p=75th percentile, std=Standard deviation.}
\label{tbl-dataset-details}
\resizebox{1\linewidth}{!}{%
\begin{tabular}{lrrrrrrrrr}
\toprule
Taxon & \# spec. & \# img. & m (mean) & m (median) & m (max) & m (min) & m (25p) & m (75p) & m (std) \\
\midrule
\midrule
\multicolumn{10}{c}{Order dataset}\\
\midrule
Araneae & 208 & 19 510 & 363.0 & 159.0 & 7 067.0 & 4.0 & 67.0 & 320.0 & 607.8 \\
Coleoptera & 204 & 17 440 & 673.5 & 213.0 & 26 013.0 & 19.0 & 67.0 & 479.0 & 2 025.1 \\
Diptera & 192 & 27 055 & 165.0 & 58.0 & 1 675.0 & 4.0 & 24.0 & 164.0 & 259.7 \\
Hemiptera & 175 & 17 915 & 365.3 & 80.0 & 20 695.0 & 9.0 & 27.0 & 345.0 & 1 045.3 \\
Hymenoptera & 201 & 26 852 & 211.3 & 93.0 & 3 474.0 & 2.0 & 30.0 & 175.0 & 406.5 \\
\midrule
All orders & 980 & 108 772 & 326.5 & 102.0 & 26 013.0 & 2.0 & 36.0 & 265.0 & 995.4 \\
\midrule
\multicolumn{10}{c}{Species dataset}\\
\midrule
Asellus aquaticus & 46 & 2 020 & 816.7 & 527.5 & 2 349.0 & 22.5 & 153.5 & 1 540.5 & 766.8 \\
Caenis horaria & 43 & 5 017 & 74.9 & 53.0 & 361.0 & 17.5 & 33.0 & 94.0 & 59.7 \\
Kageronia fuscogrisea & 47 & 3 092 & 195.2 & 123.0 & 678.5 & 43.5 & 80.0 & 289.0 & 159.2 \\
\midrule
All species & 136 & 10 129 & 259.5 & 94.0 & 2 349.0 & 17.5 & 48.0 & 215.5 & 454.7 \\
\midrule
Both datasets & 1 116 & 118 901 & 320.8 & 101.0 & 26 013.0 & 2.0 & 37.0 & 258.0 & 961.4 \\
\bottomrule
\end{tabular}}%
\end{table}

\begin{table}
    \centering
    \caption{Comparison of our dataset with previous similar invertebrate biomass datasets in the literature.}
    \label{tbl-dataset-comparison}
    \resizebox{\linewidth}{!}{
    \begin{tabular}{lrrrr}
        \toprule
        Dataset & \# of specimens & \# of weighed specimens & \% weighed \\
        \midrule
        \citet{arje2020Automatic} & 533 & 65 & 12.2 \\
        \citet{schneider2022Bulk} & 13 059 & 68 & 0.5 \\
        \citet{rehsen2026Improving} & 743 & 743 & 100.0 \\
        This study & 1 116 & 1 116 & 100.0 \\
        \bottomrule
    \end{tabular}}
\end{table}

\clearpage
\subsection{Evaluation metrics}
\label{sec-evaluation} 

We propose metric recommendations for model comparison purposes, since this is a topic not well defined in the literature.
In general, there are many different regression model metrics for different goals.
The choice of metric always introduces some bias, impacting model evaluation, and thus it is important to understand the implications of different metrics. 

The distinction between percentage errors and absolute errors for model evaluation is important, especially when estimates coming from very different dry mass ranges are aggregated.
When optimizing for smaller percentage error, the biomass of each individual specimen is estimated as correctly as possible, as the approach penalizes mistakes relative to the specimen's mass.
A \(0.010\) $mg$ error is significantly more severe if the specimen weighs \(0.005\) $mg$ (a 200\% error) compared to a specimen that weighs \(15.000\) $mg$ (a 0.0006\% error).
However, using percentage errors can lead to a larger total error, because larger \textit{absolute} errors are permitted for heavier specimens.

When optimizing for absolute error, large mistakes are not as likely to happen, making dry mass estimates for entire bulk samples more reliable.
The downside is that now larger \textit{percentage} errors are permitted for smaller specimens.
In this study, we consider both percentage and absolute errors to show a full picture of the performance of each model and approach.

For percentage errors, we use mean absolute percentage error (MAPE) and median absolute percentage error (MdAPE).
With $n$ denoting the number of specimens, specimen $i$ having the true mass of $y_i$ and the model predicting a mass of $\hat{y_i}$ for it, the equations for these metrics are
\begin{equation}\label{eq-mape}
\text{MAPE} = \frac{1}{n}\sum_{i=1}^{n}\left|\frac{y_i-\hat{y}_i}{y_i} \right|,
\end{equation}
\begin{equation}\label{eq-mdape}
\text{MdAPE} = \text{Median}\left(\left|\frac{y_i-\hat{y}_i}{y_i} \right|\right).
\end{equation}

For absolute errors, we use mean absolute error (MAE) and root mean square error (RMSE). The equations for these are
\begin{equation}\label{eq-mae}
\text{MAE} = \frac{1}{n}\sum_{i=1}^{n}\left|y_i-\hat{y}_i\right|,
\end{equation}
\begin{equation}\label{eq-rmse}
\text{RMSE} = \sqrt{\frac{1}{n}\sum_{i=1}^{n}\left(y_i-\hat{y}_i\right)^2}.
\end{equation}

In addition to the metrics above, we also used the coefficient of determination ($R^2$), a commonly used statistical metric.
$R^2$ is defined as
\begin{equation}
\text{R}^2 = 1 - \frac{\sum_{i=1}^{n}\left(y_i-\hat{y}_i\right)^2}{\sum_{i=1}^{n}\left(y_i-\bar{y}\right)^2},
\end{equation}
where $\bar{y}$ denotes the mean of the true values.
A perfect model always predicting the true values, will have $R^2$ = 1, whereas a baseline model always predicting the mean of the values has an $R^2$ = 0. We use $R^2$ on the log-transformed values.

\subsection{Biomass prediction}
\subsubsection{Convolutional neural networks}
All models are based on a simple regression task of finding a function \(f: \mathcal{X} \to \mathbb{R}^+\) that maps a natural image \(\mathbf{x} \in \mathcal{X}\) to a positive real number output.
For convolutional neural network (CNN) models, the function \(f_{\theta}\) is a neural network with parameters \(\mathbf{\theta}\), which we learn during an optimization process.
The network consists of a feature encoder $g$ (e.g., ResNet18) and a projection head $h$, s.t. $f_{\theta}(\mathbf{x}) = h(g(\mathbf{x}))$.
Given an image \(\mathbf{x}\) that represents a specimen with a dry mass \(y \in \mathbb{R}^+\), we wish to find parameters \(\theta\) so that the output \(\hat{y} = f_{\theta}(\mathbf{x})\) is as close as possible to the true value of \(y\).
The basic CNN approach trains all of the parameters in both the feature encoder and the projection head.
We also train fine-tuned models, where the feature encoder parameters are frozen, while only the projection head parameters are trained.

The multi-view CNN uses two feature encoder networks with different parameters, $g_{\theta}$ and $g_{\phi}$ that receive inputs $\mathbf{x}_1$  and $\mathbf{x}_2$ which are two images of the same specimen. The encoded features $\mathbf{z} = g(\mathbf{x})$ from both encoders are concatenated together and passed to a final projection head, producing the final output $\hat{y} = h(\text{concat}(\mathbf{z}_1, \mathbf{z}_2))$.

The metadata-aware \ac{CNN} uses metadata in addition to the images themselves. The metadata can come from the image, e.g., specimen area calculated with traditional image processing approaches, or from the sequence, e.g., the falling speed. For sequence metadata, the same value is applied to all images in the sequence. The metadata $v$ is passed to a small encoder network $\mu$ that projects the raw metadata values into an abstract representation $z_v$. This vector is then concatenated with the image feature vector before passing the value to the final projection head.

\begin{figure}
\centering
\includegraphics[width=0.7\textwidth]{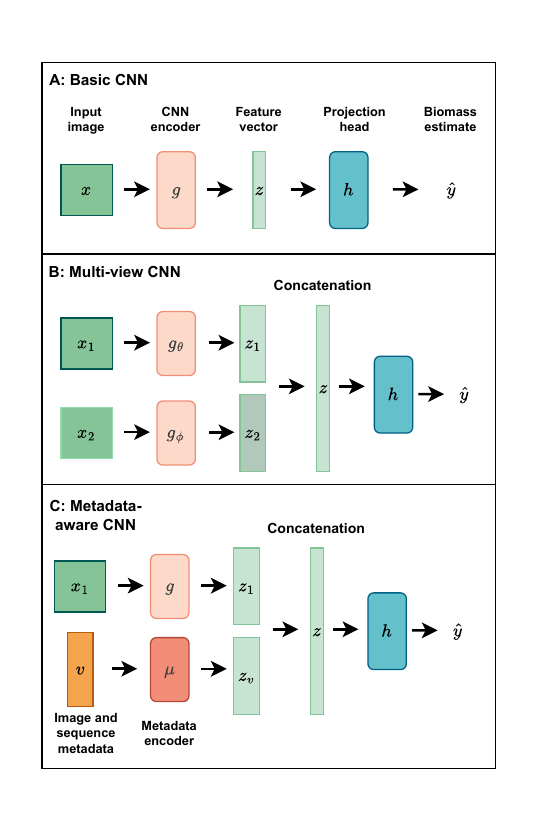}
\caption{Overview of the different CNN models used in this paper. Image inputs $x$ are passed to CNN encoders ($g$), while possible metadata $v$ is passed to a fully connected metadata encoder $\mu$. Encoders output intermediate feature vectors $z$. For the multi-view and metadata-aware architectures, feature vectors are concatenated before passing the features to a fully connected projection head ($h$) with either one or two layers. The output of this network is the final biomass estimate $\hat{y}$.}
\label{fig-modeling}
\end{figure}

An overview of different \ac{CNN} approaches can be seen in Figure~\ref{fig-modeling}. Details on \ac{CNN} training are provided in Section~\ref{sec-experimental-setup}.

\subsubsection{Linear models}
For the linear models, we fitted ordinary least squares linear models using features derived from the image sequences and their associated metadata.
Both intercept and slope parameters were estimated.
We used two features: the sinking speed of the specimen and the mean visible area of the specimen in pixels.
The use of sinking speed as a feature is a novel approach.
Previous studies have only considered the total area of images, but thanks to the image sequence information we are able to calculate a sinking speed for each specimen.

When the BIODISCOVER device captures a sequence, it images the full cuvette area (green in Figure~\ref{fig-biodiscover-overview}), but saves only crops around the specimen (blue in Figure~\ref{fig-biodiscover-overview}, on right). These crops have the information of the position of the crop in relation to the full cuvette, with information on the bounding box top, bottom, left, and right border positions.
Using the number of images \(n\) and the top border positions of the first (\(x_{max}\)) and last (\(x_{min}\)) image in the sequence, it is possible to calculate the traveled distance. Because the frame rate of the camera is fixed, the number of images \(n\) is proportional to the time that the specimen takes to fall through the cuvette. With this information it is possible to calculate the sinking speed:

\begin{equation}
s = \frac{x_{max} - x_{min}}{n},
\end{equation}
where \(s\) is the sinking speed used as a feature for the specimen.
We fitted two models: a model that uses only area as a predictor, and a model that uses both area and sinking speed as a predictor.

The biomass of a specimen was predicted separately for each image.
We used a \textit{trimmed mean} approach to determine a biomass estimate across all images of a specimen. The largest and smallest 5\% of the estimates were first removed, and the arithmetic mean was calculated for the rest of the estimates. This is conceptually close to the more commonly used trimmed mean method for aggregation.
We noticed that the trimmed mean gave better final estimates than just taking the median or mean aggregate.
The mean can be heavily influenced by a single mistaken prediction that can sometimes be very different from the other predictions.
We also performed some experiments by weighting the predictions by the visible area of the specimen, but the impact on final results was negligible.

\subsection{Experimental setup}\label{sec-experimental-setup}

We performed four different groups of experiments: (1) CNN model optimization, (2) Overall biomass estimation with CNNs and linear models, (3) Out-of-distribution generalization, and (4) End-to-end estimation with automatic classification to taxonomic groups.

\subsubsection{CNN model optimization}
The number of hyperparameters involved in CNN model training leads to a large search space for different training setups. Due to a lack of previous literature on training deep neural networks for image-based biomass estimation, we report experiments on three design choices: loss function, image augmentations, and model architecture.

First, we compared three different loss functions, L1 (absolute error), L2 (squared absolute error), and percentage error in both linear and logarithmic spaces.
For the \textit{log} models, the targets were first transformed to log-space simply by \(y_{log} = \log_e{y}\), where \(y\) is the true biomass and \(y_{log}\) is the log-transformed biomass.
The loss of the \textit{log} models was calculated on the transformed values.
Note that L1 and L2 loss functions correspond to minimizing \ac{MAE} and \ac{MSE} over each minibatch, whereas minimizing the absolute percentage error leads to minimizing \ac{MAPE}.

Next, we compared three different augmentation approaches: 1) flips and 90-degree rotations, 2) continuous rotations, and 3) a modified version of the TrivialAugment \citep{muller2021TrivialAugment} method, where we removed the geometric transformations that warp the image.
The TrivialAugment augmentation adds random brightness, color, contrast, sharpness, posterize, solarize, and equalization augmentations, with a randomly sampled strength.
We made initial small-scale experiments, which indicated that augmentations that warp the image (shearing, affine transforms, random cropping) cause the size and scale information between images to be lost, negatively affecting the model performance.


All above-listed experiments used the ResNet18 architecture \citep{he2016Deep} with a single linear projection layer as the projection head.
In the last optimization phase, we compared it to larger models.
The EfficientNet-family of models \citep{tan2019EfficientNet} was chosen due to its high performance in classification tasks even with low amounts of training data.
We trained two variants of EfficientNet, B0 and B2, to find out whether scaling the network increases performance.
As scaling did not increase performance, we opted out of training larger ResNet models.

We also trained a multi-view CNN since we have images from two cameras. We used ResNet18 models as feature encoders, and a single linear projection layer as the final projection head. We trained two versions of the multi-view model: one with log-L2 loss and one with log-L1 loss. Both losses were tested because they performed close to each other in the single-view case. The multi-view model used the flips+90\textdegree ~rotation augmentations, which were the best-performing approach for the single-view ResNet18 model.

The metadata-aware model used a small fully-connected two-layer network as the metadata encoder. The hidden layer and output layer size is 4, twice the number of input features. A rectified linear unit (ReLU) activation is added in between. For the final estimation head, that combines the image encoder CNN and the metadata encoder, we tested both single-layer and two-layer projection heads. We found that the metadata-aware model worked satisfactorily only with the two-layer projection head, with a hidden layer of 512 neurons and a ReLU activation in between. With a single-layer projection head the metadata-aware performance was significantly worse than with the image-only CNN. In order to properly compare the method to the previous methods, we also trained a image-only model with the same two-layer projection head. Again, other training parameters, such augmentations, were the same as above.

The metadata-aware model used both area and sinking speed as metadata, based on the experience from the linear models. We used two different area values: the area for the specific input image, and the mean area of the specimen's image sequence. Similarly, the sinking speed for each specimen was calculated from the full image sequence from which the specimen comes.

\subsubsection{Biomass estimation}

After \ac{CNN} model optimization, we chose two models for the fine-tuning experiment and generalization experiment on the Species dataset: ResNet18 trained with log-L1 loss and only flips+90\textdegree ~rotation augmentations, and the metadata-aware version of the same ResNet.
For linear models, the area-only model was used as a baseline, and the area+speed model and CNN models were compared to it.

We evaluated the Order dataset as a whole, without building separate models for each order.
This simulates a situation where biomass estimation is done for a heterogeneous sample of specimens and a single model is used.
The Species dataset illustrates performance on more homogeneous groups with a separate model for each taxon.

The Order dataset is significantly larger (980 specimens) than the Species datasets (46, 43, and 47 specimens, total of 136).
To simulate a situation where we would have a pretrained model available and would like to build a custom model for a single taxon, we fine-tuned the Order model from above for each taxon in the Species dataset.
We fine-tuned two versions: a fully fine-tuned model, where all parameters are trainable, and a frozen fine-tuned model, where only the projection head is trainable.
For the metadata-aware model, both the convolutional encoder and the metadata encoder were frozen.
Three fine-tuned models were trained, with the three taxa from the Species dataset.

\subsubsection{Out-of-domain biomass estimation}

Generalization to unseen, out-of-distribution groups is an important property of most predictive models.
We evaluated both CNN and linear models with out-of-distribution data.
Generalization performance on the Species dataset was done by using the best Order dataset model and simply evaluating it on each of the three Species dataset taxa.
Generalization was also tested on the taxonomic groups in the Order dataset separately, by training five new models.
Each model was trained with 4/5 of the orders, with all specimens from a single order left out as the test set.
For example, the ``Araneae'' model was trained with Coleoptera, Diptera, Hemiptera and Hymenoptera samples, and tested with Araneae - specimens from an order the model has never seen.
This was done for both CNN and linear models.
The CNN models were trained using the same protocol as the best final CNN model from the optimization experiment.

\subsubsection{Classification and end-to-end estimation}
\label{sec-classification-methods}

Estimating biomass by taxonomic groups allows for more accurate biodiversity assessments than just simply weighing the specimens in bulk.
Several other studies have shown that computer vision methods can identify many taxonomic groups reliably, given enough training data \citep{arje2020Automatic, wuhrl2024Entomoscope, deschaetzen2023Riverine, bjerge2023Accurate}.
We confirmed this using the Order dataset, by training a simple classifier that predicts the order of an unseen specimen.
We used a similar training protocol as with biomass models, using EfficientNet-B0 as the backbone, and the flips+90\textdegree ~rotation augmentations.

We used the classification model to simulate a real-world estimation setting, where the system knows nothing about a specimens' taxonomic groups or masses beforehand. The system both classifies specimens to taxonomic groups and predicts their biomasses. The model should produce dry mass distributions that resemble the true distributions of each taxonomic group. To test this system as an end-to-end pipeline, we included misclassified specimens in the final mass distributions. To evaluate performance, we used a two-sampled Kolmogorov-Smirnov test to compare the predicted dry mass distributions against the true masses.
As a complementary analysis, we also calculated Pearson's correlation coefficients within each group.

\subsubsection{Training setup}

Before training/fitting our models, we split the datasets into train, test, and validation sets.
The validation set was used for hyperparameter comparisons and model selection, while the test set was only used to calculate the final results.
We used five-fold cross-validation, splitting the full dataset to five non-overlapping test sets which together equalled the full dataset.
The splits for training, validation, and testing specimens were done with a proportion of 64\%, 16\%, and 20\% of the dataset.
As there were several images of a single specimen, we did the splitting on specimen level. This ensured that images from a single individual did not leak across sets, but all images of a specimen were in one of the three sets, ensuring that the specimens used in testing were completely unseen to the model. Splits were also stratified by taxonomic group in the Order dataset.

For all experiments, we trained five different models with the train data of one of the cross-validation folds. During evaluation, the model was used to predict the dry masses of the specimens in the corresponding test set. These test set predictions were then combined for a jackknife-style aggregate. This allowed us to estimate the final performance metrics with the full dataset, reducing variance for the metric estimates.
Confidence intervals were calculated from the this test set aggregate by bootstrapping, a method where values are repeatedly sampled with replacement and a metric is calculated from each sample.
We performed 1000 bootstrap samples to calculate confidence intervals.
The classification model was trained with the same splits and cross-validation folds as the biomass estimation models.

For CNN training, we used a batch size of 512 and the AdamW optimizer \citep{loshchilov2019Decoupled} for all experiments.
All models were initialized with the ImageNet weights.
All models in all experiments were trained for 200 epochs, i.e., passes through the full training set.
This does not mean that all models were trained for the same time: 200 epochs corresponds to 13 600 steps on the Order dataset, and 600, 800, and 1400 for \textit{Asellus aquaticus}, \textit{Kageronia fuscogrisea}, and \textit{Caenis horaria}, respectively, in the Species dataset. A step means a single pass of a batch through the neural network and updating the neural network weights via backpropagation.

After each epoch, we evaluated the model on the validation set. The best performing model in terms of loss was chosen for final evaluation with the test set. For many models, the best validation performance was achieved in the first third of training. The metadata-aware model responded the best for longer training times and the best-performing model was found usually during the last 50 epochs.

Our implementations used square input images of size \(224 \times 224\), except for EfficientNet-B2 that used an input size of $260 \times 260$.
As all images were originally square images of size $464 \times 464$, resizing the images did not warp them or change the aspect ratio.

We found that the optimization process was sensitive to small changes in the learning rate.
Therefore, we ran the \texttt{LearningRateFinder} function of the PyTorch Lightning framework with five different random seeds for 10 epochs, and chose the best learning rate separately for each loss function.
The same learning rate was then used for all later experiments.
We used a cosine annealing learning rate scheduler \citep{loshchilov2017SGDR} that decreases the learning rate based on a cosine function.
We chose to use the learning rate scheduler after small-scale experiments that showed a slight improvement compared to a constant learning rate.
The classification model was trained with a constant learning rate of 0.001.

All code was implemented in PyTorch using the Lightning framework \citep{falcon2019PyTorch} and the Scikit-learn-library \citep{scikit-learn}.
We ran our experiments concurrently in a computing cluster provided by CSC – IT Center for Science, Finland, using Nvidia Volta V100 GPUs.
However, model training can also be run in a reasonable time (< 6 hours) with a single consumer-grade GPU.
Fine-tuning with the smaller \textit{Species} datasets can be done in about 30 minutes with the V100 GPU.
More technical details on the programming environment and software can be found in the supplementary code repository.

\section{Results}\label{sec-results}

\subsection{CNN model optimization}
\label{sec-optimizationresults}

\begin{figure}[ht]
\centering\includegraphics[width=0.7\textwidth]{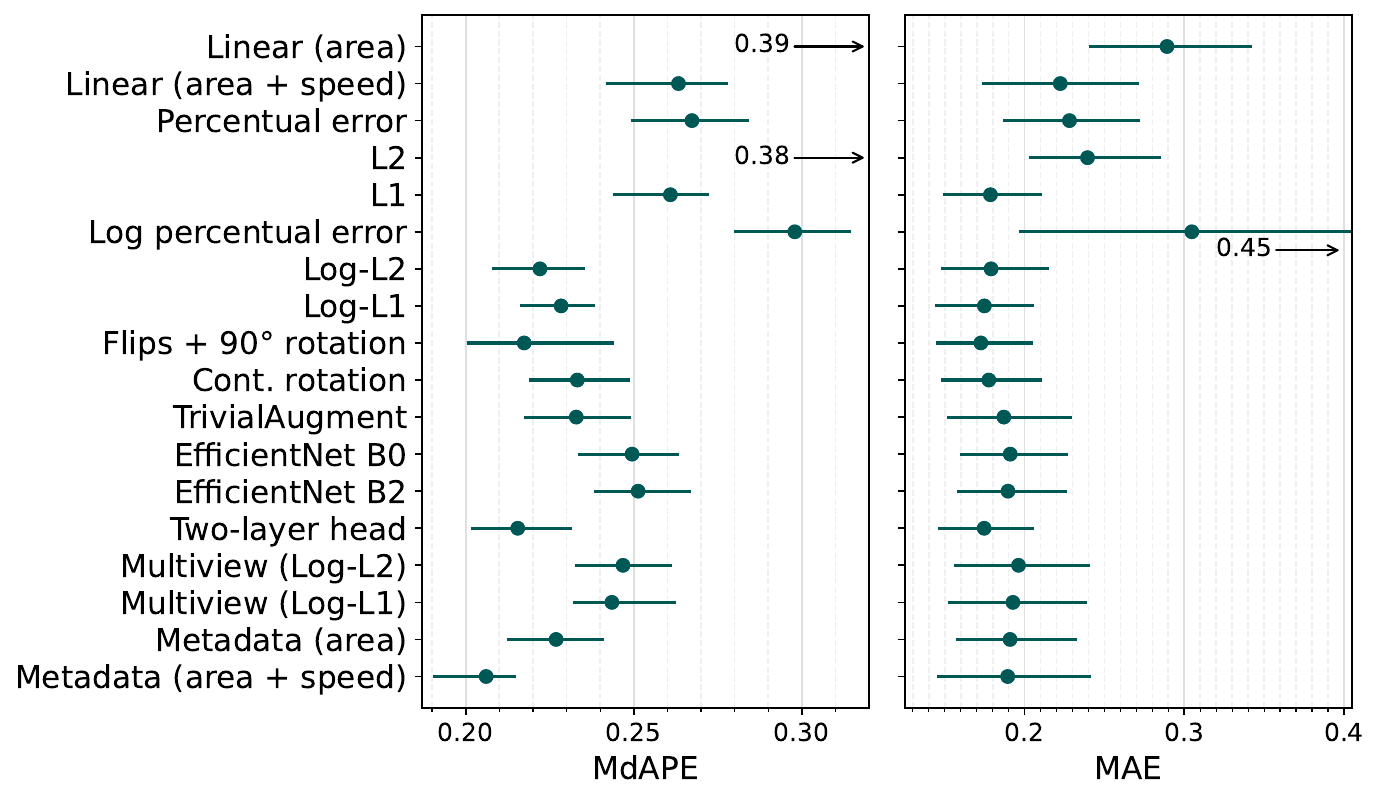}
\caption{\label{fig-results-fullcomparison}Comparison of all trained
models with bootstrapped 95\% confidence intervals.}
\end{figure}

The performance of different training setups on the Order dataset is given in Table~\ref{tbl-all-table} using five different performance metrics (described in Section~\ref{sec-evaluation}).
A subset of this table is illustrated in Figure~\ref{fig-results-fullcomparison}.

\begin{table}[ht]
\centering
\caption{Full results of model optimization experiments. The overall best performing model was the metadata-aware ResNet18 trained with log-L1 loss, flips, and 90° rotations. Best performance of each group is \underline{underlined} for each metric. Best overall performance for each metric is marked with \textbf{bold}. Bootstrapped standard deviations are reported.}
\label{tbl-all-table}
\resizebox{\linewidth}{!}{%
\begin{tabular}{lrrrrr}
\toprule
& \multicolumn{2}{c}{Absolute} & \multicolumn{2}{c}{Percentual}\\
\cmidrule(l{3pt}r{3pt}){2-3} \cmidrule(l{3pt}r{3pt}){4-5}
\textbf{Name} & \textbf{RMSE} & \textbf{MAE} & \textbf{MAPE} & \textbf{MdAPE}& \textbf{R2} \\
\midrule
\multicolumn{6}{c}{\textbf{Linear models}}\\
\midrule
Linear (area) & 1.094 ± 0.21 & 0.289 ± 0.03 & 0.578 ± 0.02 & 0.394 ± 0.01 & 0.820 ± 0.01 \\
Linear (area + speed) & \underline{0.950 ± 0.20} & \underline{0.222 ± 0.03} & \underline{0.341 ± 0.01} & \underline{0.263 ± 0.01} & \underline{0.921 ± 0.01} \\
\midrule
\multicolumn{6}{c}{\textbf{Loss functions}}\\
\midrule
Percentual error & 0.871 ± 0.14 & 0.228 ± 0.03 & 0.331 ± 0.01 & 0.267 ± 0.01 & 0.906 ± 0.01 \\
L2               & 0.816 ± 0.18 & 0.239 ± 0.03 & 0.673 ± 0.03 & 0.378 ± 0.02 & 0.000 ± 0.00 \\
L1               & 0.624 ± 0.13 & 0.179 ± 0.02 & 0.409 ± 0.02 & 0.261 ± 0.01 & 0.906 ± 0.01 \\
Log percentual error & 2.491 ± 1.11 & 0.305 ± 0.08 & 0.452 ± 0.02 & 0.298 ± 0.01 & 0.886 ± 0.01 \\
Log-L2           & 0.665 ± 0.15 & 0.179 ± 0.02 & \underline{0.321 ± 0.01} & \underline{0.222 ± 0.01} & 0.926 ± 0.00 \\
Log-L1           & \underline{0.612 ± 0.12} & \underline{0.175 ± 0.02} & 0.326 ± 0.01 & 0.228 ± 0.01 & \underline{0.927 ± 0.00} \\
\midrule
\multicolumn{6}{c}{\textbf{Augmentation}}\\
\midrule
Flips + 90° rotation & \underline{0.618 ± 0.13} & \underline{\textbf{0.173 ± 0.02}} & 0.328 ± 0.01 & \underline{0.217 ± 0.01} & 0.928 ± 0.00 \\
Cont. rotation & 0.636 ± 0.12 & 0.178 ± 0.02 & \underline{0.316 ± 0.01} & 0.233 ± 0.01 & \underline{0.929 ± 0.00} \\
TrivialAugment & 0.742 ± 0.18 & 0.187 ± 0.02 & 0.360 ± 0.01 & 0.233 ± 0.01 & 0.921 ± 0.01 \\
\midrule
\multicolumn{6}{c}{\textbf{Other architectures and approaches}}\\
\midrule
EfficientNet B0 & 0.655 ± 0.13 & 0.191 ± 0.02 & 0.348 ± 0.01 & 0.249 ± 0.01 & 0.921 ± 0.01 \\
EfficientNet B2 & 0.656 ± 0.13 & 0.190 ± 0.02 & 0.354 ± 0.01 & 0.251 ± 0.01 & 0.919 ± 0.01 \\
Two-layer head & \underline{\textbf{0.610 ± 0.11}} & \underline{0.175 ± 0.02} & 0.335 ± 0.01 & 0.215 ± 0.01 & 0.925 ± 0.00 \\
Multi-view (Log-L2) & 0.841 ± 0.22 & 0.196 ± 0.03 & 0.346 ± 0.01 & 0.247 ± 0.01 & 0.922 ± 0.00 \\
Multi-view (Log-L1) & 0.863 ± 0.28 & 0.193 ± 0.03 & 0.366 ± 0.01 & 0.243 ± 0.01 & 0.919 ± 0.01 \\
Metadata (area) & 0.720 ± 0.17 & 0.191 ± 0.02 & 0.346 ± 0.01 & 0.227 ± 0.01 & 0.921 ± 0.01 \\
Metadata (area + speed) & 0.966 ± 0.24 & 0.189 ± 0.03 & \underline{\textbf{0.290 ± 0.01}} & \underline{\textbf{0.206 ± 0.01}} & \underline{\textbf{0.941 ± 0.00}} \\
\bottomrule
\end{tabular}}

\end{table}

Log-L1 and log-L2 losses were the best-performing loss functions and performed similar to each other.
Linear L1 loss was close to these results in absolute metrics, but log-L1 loss achieved better percentage error results.
Log percentage error was the worst-performing loss function and did not produce good performance even on percentage error metrics.
Due to overall performance, log-L1 was used in the remaining model optimization experiments.

Complicated augmentations and increasing model size did not improve performance.
90 degree rotations performed slightly better than continuous rotations.
Additional augmentations from TrivialAugment did not make the models perform better.
Also, neither larger EfficientNet models nor using a multi-view model with two ResNet18 encoders improved performance.
The multi-view model trained with log-L2 loss performed better than the one with log-L1 loss in terms of some metrics.

Figure~\ref{fig-results-fullcomparison} shows that the small differences in augmentations and model architectures are often within the 95\% confidence interval calculated by bootstrapping the results.
Overall, the best model was the metadata-aware CNN. Measured in absolute metrics, the single-view model with a two-layer projection head being is slightly better than the metadata-aware CNN.

\subsection{Biomass estimation}\label{sec-biomass-estimation}

The overall results for biomass estimation with both CNNs and linear models can be seen in Table~\ref{tbl-overall-biomass-estimation}, illustrating performance in three different scenarios. In \emph{in-distribution} training, the models were trained and evaluated with the same dataset (Order/Species). In \emph{fine-tuning}, the \ac{CNN} models were first trained with the Order dataset, then fine-tuned with the Species dataset, and finally evaluated on the Species dataset. In \emph{out-of-distribution evaluation}, the models are trained only with the Order dataset, without further fine-tuning, and evaluated on the Species dataset.

\begin{table}[ht]
\caption{Biomass estimation results. \textit{Lin. A} = Linear model using only area as predictor. \textit{Lin. A+S} = Linear model with area and speed predictors. \textit{CNN} = CNN model trained with in-domain data. \textit{CNN A+S} = Metadata-aware CNN with area+speed predictors. \textit{FT CNN} = CNN model pretrained with OOD data and fine-tuned with in-domain data. \textit{FT CNN A+S} = Similarly fine-tuned metadata-aware CNN model. \textit{FT CNN Frz} = FT model with convolutional layers frozen during training. \textit{FT CNN A+S Frz} = Metadata-aware model with CNN encoder and metadata encoder frozen. \textit{OOD} = CNN model trained with out-of-domain data. \textit{OOD A+S} = Metadata-aware CNN model trained with out-of-domain data. \textit{Lin. OOD} = Linear A+S model trained with OOD data. Best models for each group are marked with \textbf{bold}. Bootstrapped standard deviations are reported.}
\label{tbl-overall-biomass-estimation}
\centering
\resizebox{\linewidth}{!}{%
\begin{tabular}{lcccccccc}
\toprule
 &  \multicolumn{2}{c}{Order Dataset} & \multicolumn{6}{c}{Species Dataset}\\
\cmidrule(l{3pt}r{3pt}){2-3} \cmidrule(l{3pt}r{3pt}){4-9}
 & & & \multicolumn{2}{c}{Asellus aquaticus} & \multicolumn{2}{c}{Caenis horaria} & \multicolumn{2}{c}{Kageronia fuscogrisea} \\
 \cmidrule(l{3pt}r{3pt}){4-5} \cmidrule(l{3pt}r{3pt}){6-7} \cmidrule(l{3pt}r{3pt}){8-9}
Method  & MAE & MdAPE & MAE & MdAPE & MAE & MdAPE & MAE & MdAPE \\
\midrule
\multicolumn{9}{c}{\textbf{In-distribution evaluation}}\\
\midrule
Lin. A (Baseline)  & 0.289 ± 0.03 & 0.394 ± 0.01 & 0.144 ± 0.02 & \textbf{0.116 ± 0.03} & 0.035 ± 0.01 & 0.403 ± 0.06 & \textbf{0.070 ± 0.01} & \textbf{0.209 ± 0.04} \\
Lin. A+S  & 0.222 ± 0.03 & 0.263 ± 0.01 & \textbf{0.126 ± 0.02} & 0.126 ± 0.02 & \textbf{0.022 ± 0.00} & \textbf{0.234 ± 0.03} & 0.079 ± 0.02 & 0.226 ± 0.04 \\
CNN  & \textbf{0.173 ± 0.02} & 0.217 ± 0.01 & 0.198 ± 0.03 & 0.202 ± 0.04 & 0.038 ± 0.01 & 0.324 ± 0.05 & 0.082 ± 0.01 & 0.302 ± 0.05 \\
CNN+A+S  & 0.189 ± 0.03 & \textbf{0.206 ± 0.01}
 & 0.366 ± 0.03 & 0.357 ± 0.12 & 0.070 ± 0.01 & 0.994 ± 0.28 & 0.137 ± 0.01 & 0.704 ± 0.23 \\
\midrule
\multicolumn{9}{c}{\textbf{Fine-tuning evaluation}}\\
\midrule
FT CNN  & - & - & 0.212 ± 0.03 & 0.181 ± 0.02 & 0.024 ± 0.00 & 0.255 ± 0.04 & \textbf{0.061 ± 0.01} & \textbf{0.186 ± 0.03} \\
FT CNN Frz  & - & - & \textbf{0.187 ± 0.03} & \textbf{0.152 ± 0.02} & 0.026 ± 0.00 & 0.249 ± 0.05 & 0.072 ± 0.01 & 0.218 ± 0.02 \\
FT CNN A+S  & - & - & 0.291 ± 0.04 & 0.320 ± 0.03 & \textbf{0.023 ± 0.00} &\textbf{ 0.231 ± 0.03} & 0.177 ± 0.05 & 0.216 ± 0.04 \\
FT CNN A+S Frz & - & - & 0.402 ± 0.05 & 0.448 ± 0.03 & 0.027 ± 0.00 & 0.261 ± 0.05 & 0.216 ± 0.06 & 0.261 ± 0.04 \\
\midrule
\multicolumn{9}{c}{\textbf{Out-of-distribution evaluation}}\\
\midrule
OOD Lin. A+S & \multicolumn{2}{c}{See Table~\ref{tbl-generalization}} & \textbf{0.302 ± 0.04} & \textbf{0.242 ± 0.04} & \textbf{0.024 ± 0.00} & \textbf{0.262 ± 0.03} & \textbf{0.156 ± 0.05} & \textbf{0.250 ± 0.07 }\\
OOD CNN  & \multicolumn{2}{c}{See Table~\ref{tbl-generalization}}  & 2.212 ± 0.14 & 2.552 ± 0.41 & 1.375 ± 0.05 & 20.451 ± 3.24 & 1.771 ± 0.13 & 8.150 ± 1.01 \\
OOD CNN A+S &  \multicolumn{2}{c}{See Table~\ref{tbl-generalization}} & 1.389 ± 0.24 & 1.389 ± 0.39 & 1.988 ± 0.27 & 18.833 ± 2.52 & 2.768 ± 0.54 & 7.494 ± 1.35 \\
\bottomrule
\end{tabular}}
\end{table}

\begin{figure}
\centering
\includegraphics[width=\textwidth]{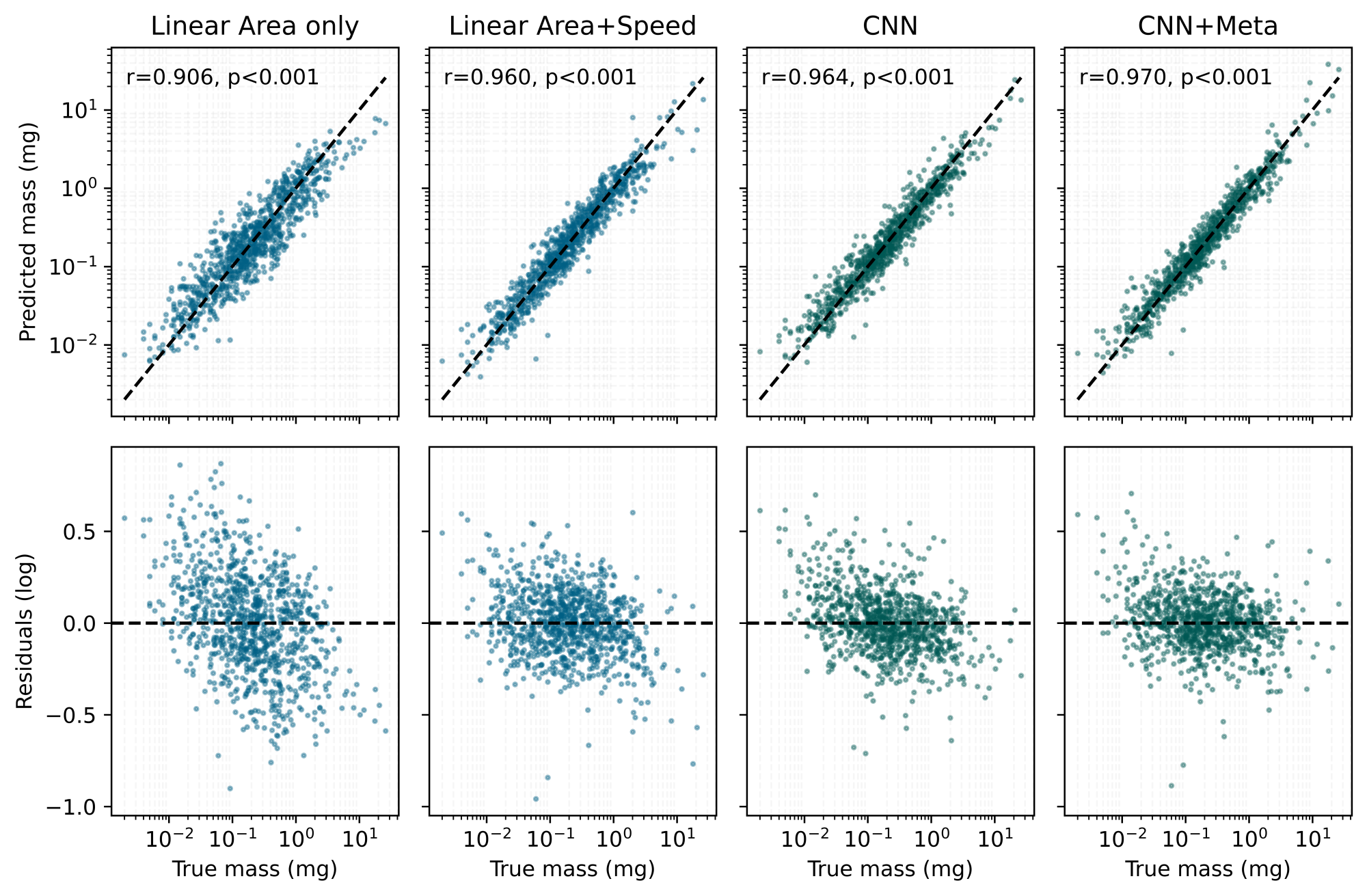}
  \caption{Distribution of log-transformed dry mass predictions against true values, with residuals calculated on the log-transformed scale. Pearson's correlation coefficients (r) are shown for each method. 1:1 line is shown for reference. All models overestimate small samples and underestimate larger samples. }
  \label{fig-regression}
\end{figure}

The results show that the performance varies across different models and metrics. With the larger and more heterogeneous Order dataset, the CNN models outperform the linear models.
On the smaller Species dataset, linear models perform better.
Pretraining CNNs with a larger dataset generally improves performance in the smaller datasets, as seen by comparing the in-distribution CNNs to the fine-tuned CNNs.
For basic CNN, freezing the convolutional layers and training only the final layer yields best results, while for the metadata-aware model this has an opposite effect.
The distribution and residuals of the Order dataset predictions are provided in Figure~\ref{fig-regression} for the area only model, area+speed model, basic CNN, and metadata-aware CNN.
Overall, the residuals show that generally all models overshoot the mass of lighter specimens, and undershoots the mass of heavier specimens.
All methods produce biomass estimates where true and predicted log-transformed dry-masses are strongly correlated. The highest agreement was obtained with the metadata-aware CNN model (Pearson's $r = 0.970, p < 0.001$).

\subsection{Out-of-distribution evaluation}\label{sec-generalization}

Table~\ref{tbl-overall-biomass-estimation} shows that the OOD generalization performance of the CNN model is significantly worse than with the linear model.
Metadata-aware model has better generalization performance than the basic CNN, but still falls far from the linear model.
Generalization results on the Order dataset can be seen in Table~\ref{tbl-generalization}.
For this larger dataset, the CNN models show improved generalization performance, but the linear model still shows the strongest generalization performance.
For Araneae and Hemiptera, the addition of metadata to the CNN improved performance. Overall, CNN and linear models perform close to each other, except for Coleoptera and Diptera, where the metadata-aware model is worse by a larger margin.

\begin{table}
\caption{Generalization performance on the order dataset. The models are trained on all other classes except for the one in the \textit{Outlier taxon}-column, and tested on the \textit{Outlier taxon} only. Bootstrapped standard deviations are reported.}
\label{tbl-generalization}
\centering
\resizebox{\linewidth}{!}{%
\begin{tabular}{lllllll}
\toprule
Outlier taxon & Model&  RMSE & MAE & MAPE & MdAPE & R2 \\
\midrule
\multirow[t]{2}{*}{Araneae} & Linear & 0.262 ± 0.05 & \textbf{0.120 ± 0.02} & 0.265 ± 0.02 & \textbf{0.191 ± 0.02} & \textbf{0.938 ± 0.01} \\
 & CNN & 0.450 ± 0.08 & 0.212 ± 0.03 & 0.491 ± 0.05 & 0.342 ± 0.04 & 0.840 ± 0.02 \\
 & CNN+Meta &\textbf{0.246 ± 0.05} & 0.121 ± 0.02 & \textbf{0.264 ± 0.02} & 0.204 ± 0.01 & 0.930 ± 0.01 \\
\cline{2-7}
\multirow[t]{2}{*}{Coleoptera} & Linear & 1.873 ± 0.52 & \textbf{0.508 ± 0.14} & \textbf{0.345 ± 0.02} & \textbf{0.304 ± 0.02} & \textbf{0.903 ± 0.01} \\
 & CNN & \textbf{1.683 ± 0.39} & 0.545 ± 0.12 & 0.391 ± 0.03 & 0.316 ± 0.03 & 0.882 ± 0.02 \\
 & CNN+Meta & 2.333 ± 0.60 & 0.657 ± 0.17 & 0.397 ± 0.02 & 0.359 ± 0.04 & 0.868 ± 0.01 \\
\cline{2-7}
\multirow[t]{2}{*}{Diptera} & Linear & \textbf{0.238 ± 0.02} & 0.140 ± 0.02 & \textbf{0.568 ± 0.04} & 0.404 ± 0.03 & \textbf{0.865 ± 0.02} \\
 & CNN & 0.257 ± 0.06 & \textbf{0.118 ± 0.02} & 0.660 ± 0.07 & \textbf{0.319 ± 0.04 }& 0.826 ± 0.02 \\
 & CNN+Meta & 0.330 ± 0.03 & 0.184 ± 0.02 & 0.700 ± 0.06 & 0.565 ± 0.07 & 0.821 ± 0.02 \\
\cline{2-7}
\multirow[t]{2}{*}{Hemiptera} & Linear & 0.507 ± 0.08 & 0.230 ± 0.04 & 0.339 ± 0.02 & 0.315 ± 0.02 & 0.900 ± 0.02 \\
 & CNN & 0.737 ± 0.16 & 0.304 ± 0.06 & 0.410 ± 0.03 & 0.364 ± 0.03 & 0.861 ± 0.02 \\
 & CNN+Meta &\textbf{0.411 ± 0.07} &\textbf{0.182 ± 0.03} &\textbf{0.337 ± 0.03 }&\textbf{0.273 ± 0.02} & \textbf{0.922 ± 0.01} \\
\cline{2-7}
\multirow[t]{2}{*}{Hymenoptera} & Linear & \textbf{0.505 ± 0.22} & \textbf{0.106 ± 0.04} & \textbf{0.307 ± 0.03} &\textbf{0.221 ± 0.02 }& \textbf{0.937 ± 0.01} \\
 & CNN & 0.530 ± 0.27 & 0.107 ± 0.04 & 0.463 ± 0.07 & 0.241 ± 0.02 & 0.887 ± 0.02 \\
 & CNN+Meta & 0.761 ± 0.34 & 0.172 ± 0.06 & 0.441 ± 0.05 & 0.301 ± 0.03 & 0.900 ± 0.01 \\
\bottomrule
\end{tabular}}
\end{table}

\subsection{Classification and end-to-end estimation}\label{sec-end-to-end}
\begin{figure}
\centering
\includegraphics[width=0.5\textwidth]{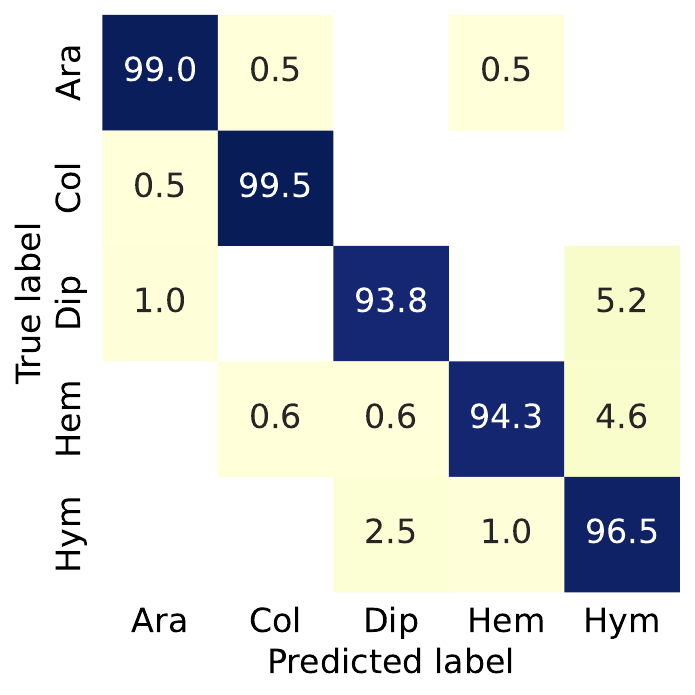}
  \caption{Classification confusion matrix, as percentages of the true class. See Table~\ref{tbl-dataset-details} for number of specimens for each class. Ara=Araneae, Col=Coleoptera, Dip=Diptera, Hem=Hemiptera, Hym=Hymenoptera}
  \label{fig-classification}
\end{figure}

Table~\ref{tab-classification} and Figure~\ref{fig-classification} show the results of the Order dataset classifier described in Section~\ref{sec-classification-methods}.
Classification performance is high across all groups, with a minimum recall score of 0.938 for Diptera and a maximum recall of 0.995 for Coleoptera.
Most common error is classifying Diptera and Hemiptera specimens as Hymenoptera.

\begin{figure}
\centering
\includegraphics[width=\textwidth]{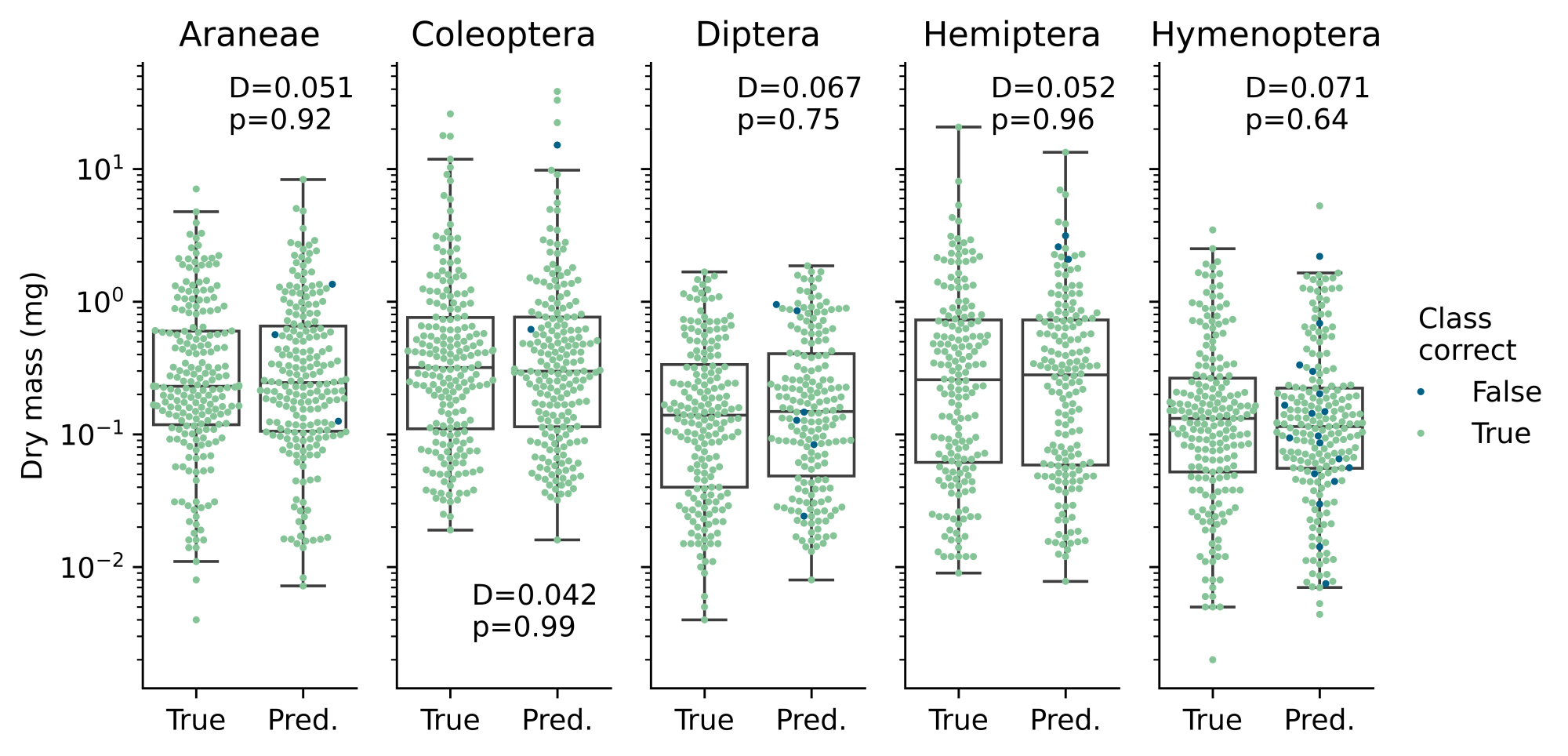}
  \caption{Swarm plot of classification combined with biomass estimation. Each specimen is represented as a dot. True dry masses are plotted next to predicted values. Specimen dry masses are plotted based on their \textit{predicted} class. Classification mistakes are highlighted with a darker color. Each dot corresponds to a single specimen. Two-sample Kolmogorov-Smirnov test statistics are reported for each group.}
  \label{fig-cls-weight}
\end{figure}

Figure ~\ref{fig-cls-weight} shows the results for the combined classification and biomass estimation pipeline.
All distributions are highly similar, with small KS distances for all groups ($D = 0.042-0.071$), and no significant differences detected. The largest difference in predicted mass distribution is in Hymenoptera, ($D = 0.071, p = 0.64$), which has the most misclassified specimens from Diptera and Hemiptera.
Pearson's correlation coefficients were high for all groups ($r = 0.958 - 0.974$, all $p < 0.001$).

\begin{table}[ht]
    \centering
    \small
    \caption{Classification metrics for the Order dataset.}
    \label{tab-classification}
\begin{tabular}{lrrr}
\toprule
 & Precision & Recall & F1-score \\
\midrule
Araneae & 0.986 & 0.990 & 0.988 \\
Coleoptera & 0.990 & 0.995 & 0.993 \\
Diptera & 0.968 & 0.938 & 0.952 \\
Hemiptera & 0.982 & 0.943 & 0.962 \\
Hymenoptera & 0.915 & 0.965 & 0.939 \\
\bottomrule
\end{tabular}
\end{table}

\clearpage

\section{Discussion}\label{sec-discussion}
\subsection{Overall results}
Our results show that mean visible area and sinking speed are strong predictors of dry mass. Sinking speed improved both the linear estimator and the metadata-aware CNN model.
Because sinking speed serves as a proxy for density, it can be combined with area measurements to estimate specimen mass. 
The end-to-end estimation system experiment indicates that community-level estimation accurately reflects the true biomass distributions for different taxonomic groups, producing more fine-grained insight to the community structure than an aggregate biomass metric.

In-domain and out-of-domain results (Tables~\ref{tbl-overall-biomass-estimation} and \ref{tbl-generalization}) show that the linear model excels on the small and homogeneous Species dataset, while the CNN performs better on the larger and heterogeneous Order dataset. The CNN struggles on the smaller dataset most likely due to overfitting. This is supported by the fact that the generalization performance of the CNN is significantly worse. Many models also achieved their lowest validation loss early in the training, with longer training times leading to worse performances, and only the metadata-aware model being more receptive for longer training times.

However, when the dataset size is increased with the Order dataset, CNN performance improves significantly.
Overall biomass estimation is better on the more heterogeneous data containing multiple taxonomic groups.
This is likely because CNNs can learn complex mappings from visual features to dry mass.
A single linear model cannot capture these complex relationships in a dataset containing very different visual groups.
We also see that out-of-distribution  performance increases, and for some groups surpasses the linear model performance.
Further research is still required to explore these CNN properties in biomass estimation.

\subsection{CNN model optimization}
Usually, heavy augmentation \citep{muller2021TrivialAugment} improves performance in deep neural networks.
However, our results indicate that in the context of biomass estimation, some augmentation methods can actually harm performance, and simple, non-warping transformations yielded the best results.
Warping augmentations, such as affine transformations and random cropping produced consistently worse results than non-warping augmentations, probably due to the fact that unlike in classification, dry mass is not invariant to the physical dimensions of the object in the image.
Heavy augmentations in TrivialAugment slightly decreased model performance, even after removing warping augmentations.
Discrete 90-degree rotations outperformed continuous rotations, probably because they avoid empty corner padding. This effect may diminish as dataset size grows and the model learns to disregard padding.

It should be noted that our training setup and hyperparameter optimization experiments were not designed to produce the best possible model, but to compare different hyperparameters in a consistent way. It is likely that better performing models can be trained with larger datasets and fine-grained hyperparameter optimization and engineering.

\subsection{Limitations}
The models have still some limitations that could be addressed in the future.
The linear model is sensitive to outliers in sinking speed, which can occur if only a few images can be captured from the specimen. For example, Figure~\ref{fig-weight-vs-images} shows that \textit{Kageronia fuscogrisea} has some outliers with only 1-2 images per specimen, causing the area-only model to outperform area+speed model in Table~\ref{tbl-overall-biomass-estimation}.
The CNN models might be further improved by incorporating additional information about the image sequences. Currently, the sequence information is used only to compute sinking speed and mean area, and when aggregating results using quantile mean.
Models that take the full sequence as input could be an useful alternative.
Additional covariates, such as total sample wet weight used to calibrate summed predictions, could be tested in the future.
Although the data did not indicate large differences in the sinking speeds of specimens stored in different ethanol concentrations (see Figure~\ref{fig-dataset-stats}, this relationship should also be studied further.

Imaging with the BIODISCOVER system still requires a considerable amount of manual handling, requiring more effort than bulk wet weight measurements but less than drying or individually measuring specimens by hand. Nevertheless, combined with automated and accurate taxonomic assignment the system can still produce accurate relative abundance information and size distributions for different groups of interest, illustrated by the end-to-end experiment in Section \ref{sec-end-to-end}.

\subsection{Recommendations}
Performance metrics of many models are within each others' confidence intervals.
This makes it harder to assess whether a single augmentation or architecture is absolutely better than the other.
We can, however make certain recommendations based on our study.
Linear models with area and sinking speed predictors work better with smaller datasets with homogenous visual features.
Metadata-aware CNNs should be used with larger datasets or when the dataset has heterogenous visual features.
When training a CNN, it can be useful to use pre-trained weights if the dataset is not sufficiently large.
The Species dataset fine-tuning performance shows that a dataset of only \textasciitilde 50 specimens per group is enough to produce sufficient models via fine-tuning.
Large-scale benchmark datasets should be collected from various taxonomic groups to thoroughly further evaluate these methods.

Log-space losses (log-L1, log-L2) seem to work better than their non-log counterparts.
When conducting biomass estimation experiments, both percentage and absolute metrics should be discussed, as they have very different interpretations.
However, optimizing for absolute error overrides mistakes in very small specimens, where percentage error can be large.
Thus, if a choice between these two must be made, percentage errors are generally more useful.
We emphasize that these recommendations are based on our limited datasets and clearly more studies should be conducted.
We encourage the collection of datasets with varying orders and large diversity and making these datasets publicly available with easy access.

\section{Conclusions}\label{sec-conclusions}
In this paper, we evaluated different strategies for estimating small arthropod biomass with computer vision methods. We have analyzed both linear models, based on features extracted from image sequences, and deep learning approaches. We proposed using specimen sinking speed as a useful predictor for dry mass. This predictor is used as input for a linear regression model and a metadata-aware convolutional neural network model. We evaluated and discussed different modeling approaches, and performed a broad range of experiments including in-distribution evaluation, fine-tuning evaluation, and out-of-distribution evaluation.

Our results show that automated biomass estimation with computer vision methods is a viable alternative to dry-weighing. The BIODISCOVER system, with the ability to record sinking speed information, can produce good biomass estimates for a broad range of taxa, with a median percentage error of 10-20\% for individuals, and accurate community-level dry mass distribution estimations for automatically classified groups.

\section{CRediT authorship contribution statement}
\noindent\textbf{Mikko Impiö:} Conceptualization, Data curation, Formal analysis, Investigation, Methodology, Project administration, Resources, Software, Validation, Visualization, Writing - original draft, Writing - review \& editing.
\textbf{Philipp M. Rehsen:} Conceptualization, Data curation, Formal analysis, Investigation, Methodology, Resources, Software, Validation, Visualization, Writing – review \& editing.
\textbf{Jarrett Blair:} Conceptualization, Investigation, Validation, Writing – review \& editing.
\textbf{Cecilie Mielec:} Data curation, Investigation, Resources, Validation, Writing – review \& editing.
\textbf{Arne J. Beermann:} Conceptualization, Resources, Supervision, Validation, Writing – review \& editing.
\textbf{Florian Leese:} Conceptualization, Resources, Supervision, Validation, Writing – review \& editing.
\textbf{Toke T. Høye:} Conceptualization, Funding acquisition, Methodology, Resources, Supervision, Validation, Writing – review  \& editing.
\textbf{Jenni Raitoharju:} Conceptualization, Funding acquisition, Methodology, Project administration, Resources, Supervision, Validation, Writing – original draft, Writing – review  \& editing.

\section{Declaration of competing interest}
The authors declare that they have no known competing financial interests or personal relationships that could have appeared to influence the work reported in this paper.

\section{Acknowledgements}
This work was supported by Research Council of Finland project 333497. This research was supported by Biodiversa+, the European Biodiversity Partnership, in the context of the DNAquaIMG project under the 2022-2023 BiodivMon joint call. It was co-funded by the European Comission (GA No. 101052342) and the following national funding organisations: Deutsche Forschungsgemeinschaft e.V. (DFG), VDI/VDE Innovation + Technik GmbH on behalf of BMBF (BMBF-VDI/VDE-IT), Suomen Akatemia (AKA), Ministry of the Environment (MoE\_FI), Agence Nationale de la Recherche (ANR), Office français de la biodiversité (OFB), Narodowe Centrum Nauki (NCN), Environmental Protection Agency of Ireland (Irish EPA), FCT - Fundação para a Ciência e a Tecnologia, I.P. in the scope of the project referenced BiodivMon/0002/2022 and DOI identifier \\ https://doi.org/10.54499/BiodivMon/0002/2022, Technology Agency of the Czech Republic (TA CR), Fonds zur Förderung der wissenschaftlichen \\ Forschung (FWF), Innovation Fund Denmark (IFD). \\ This work was supported by Jægernes Naturfond and European Union's Horizon Europe Research and Innovation programme, under Grant Agreement No. 101060639 (MAMBO). We thank CSC - IT Center for Science, Finland for computational resources.
We thank Riku Karjalainen from Syke for their assistance in collecting the ``Species'' subset of the dataset. We thank Jakob O. Kaare-Rasmussen for discussions regarding the sinking speed approach for the linear estimator. 

\section{Data availability}
The image datasets with their accompanying metadata are publicly available at https://doi.org/10.5281/zenodo.20019571. Code used for data preprocessing, model training, prediction, and analysis is available at \\
https://github.com/mikkoim/invertebrate-biomass/ enabling full reproducibility of the results.




\bibliographystyle{elsarticle-harv}\biboptions{authoryear}
\bibliography{references.bib}
\end{document}